\documentclass[10pt,journal,compsoc]{IEEEtran}

\usepackage[table,xcdraw]{xcolor} 
\usepackage{booktabs} 
\usepackage{amsmath,amsfonts}
\usepackage{algorithm}
\usepackage{algorithmic}
\usepackage{array}
\usepackage[caption=false,font=normalsize,labelfont=sf,textfont=sf]{subfig}
\usepackage{textcomp}
\usepackage{stfloats}
\usepackage{url}
\usepackage{verbatim}
\usepackage{graphicx}
\usepackage{graphicx} 
\usepackage{amsmath}
\usepackage{amsfonts}
\usepackage{multirow}
\usepackage{color,xcolor}
\usepackage{bbding}
\usepackage{graphicx}
\usepackage{amsmath}
\usepackage{amssymb}
\usepackage{booktabs}
\usepackage{enumitem}
\usepackage{multirow}

\definecolor{lightgray}{gray}{.9}
\definecolor{forestgreen}{RGB}{0,128,69}
\definecolor{cvprpink}{RGB}{219,112,147}
\definecolor{cvpryellow}{RGB}{255, 191, 0}
\definecolor{cvprblue}{rgb}{0.21,0.49,0.74}

\usepackage[colorlinks,
            linkcolor=red,       
            anchorcolor=blue,  
            citecolor=blue,        
            ]{hyperref}

\usepackage{amsthm} 

\newtheorem{definition}{Definition}

\usepackage{orcidlink}

\usepackage{bm}

\usepackage[export]{adjustbox}

\usepackage{threeparttable}
\newcolumntype{I}{!{\vrule width 1pt}}
\makeatletter
\newcommand{\thickhline}{%
    \noalign {\ifnum 0=`}\fi \hrule height 1pt
    \futurelet \reserved@a \@xhline
}
\makeatother
\usepackage{float}
\usepackage{pifont}
\usepackage{ragged2e}

\ifCLASSOPTIONcompsoc
  \usepackage[nocompress]{cite}
\else
  \usepackage{cite}
\fi
\usepackage{enumitem}
\begin{document}
\title{Calibrating Biased Distribution in VFM-derived Latent Space via Cross-Domain Geometric Consistency}
\author{Yanbiao Ma~\orcidlink{0000-0002-8472-1475},
        Wei Dai~\orcidlink{0000-0003-3354-9617},
		Bowei Liu~\orcidlink{0000-0003-3354-9617},
        Jiayi Chen~\orcidlink{0000-0003-3354-9617},
        Wenke Huang~\orcidlink{0000-0002-5669-9354},
        Guancheng Wan~\orcidlink{0000-0002-5669-9354},\\
        Zhiwu Lu~\orcidlink{0000-0001-6429-7956},~\IEEEmembership{Senior Member,~IEEE,}
        Junchi Yan~\orcidlink{0000-0001-9639-7679},~\IEEEmembership{Senior Member,~IEEE}

\thanks{Yanbiao Ma and Zhiwu Lu are with the Gaoling School of Artificial Intelligence, Renmin University of China. Bowen Liu is with Tsinghua University. Wei Dai and Jiayi Chen are with Xidian University. Wenke Huang is with Wuhan University. Junchi Yan is with Shanghai Jiao Tong University.}
\thanks{Correspondence author: Zhiwu Lu, Junchi Yan}
\thanks{E-mail: luzhiwu(@)ruc.edu.cn, yanjunchi@sjtu.edu.cn}
}

\IEEEtitleabstractindextext{%
\begin{abstract}
Despite the fast progress of deep learning, one standing challenge is the gap of the observed training samples and the underlying true distribution. There are multiple reasons for the causing of this gap e.g. sampling bias, noise etc. In the era of foundation models, we show that when leveraging the off-the-shelf (vision) foundation models (e.g., CLIP, DINOv2) for feature extraction, the geometric shapes of the resulting feature distributions exhibit remarkable transferability across domains and datasets. To verify its practical usefulness, we embody our geometric knowledge-guided distribution calibration framework in two popular and challenging settings: federated learning and long-tailed recognition. In the federated setting, we devise a technique of acquiring the global geometric shape under privacy constraints, then leverage this knowledge to generate new samples for clients, in the aim of bridging the gap between local and global observations. In long-tailed learning, it utilizes the geometric knowledge transferred from sample-rich categories to recover the true distribution for sample-scarce tail classes. Comprehensive experiments show that our proposed geometric knowledge-guided distribution calibration effectively overcomes information deficits caused by data heterogeneity and sample imbalance, with boosted performance across benchmarks.
Code published at: \url{https://github.com/WeiDai-David/2025CVPR_GGEUR}.
\end{abstract}

}

\maketitle

\IEEEdisplaynontitleabstractindextext

\IEEEpeerreviewmaketitle

\section{Introduction}
\label{sec:intro}

\IEEEPARstart{D}{espite} the fast progress of modern deep neural networks including the recent development of vision and multi-modality foundation models~\cite{CLIP,zhang2024vision,yan3}, there still exists a fundamental challenge in the scenarios with limited samples \cite{survey,tpami5}. It is often the case that the training data relied upon by models is often only a local \cite{LTdata}, sparse \cite{DSB}, and biased observation \cite{yan1} of the underlying ideal global data distribution. This distribution missing phenomenon manifests in various forms: in federated learning, it appears as label skew and domain skew due to data silos among clients \cite{huang3,fl3,karimireddy2020scaffold}, causing a severe misalignment between local data distributions and the global ideal distribution, thereby leading to divergent or even conflicting local optimization directions \cite{huang2,li2023no,mu2023fedproc}. In long-tailed recognition, it is characterized by the extreme scarcity of samples in tail classes, preventing the model from capturing the true and complete shape of their distributions \cite{FDC,tpami6}. Despite the differing scenarios, the essence is highly unified—models learn from incomplete information, lacking a comprehensive understanding of the overall structure of the real world.

Conventional solutions, such as weighting loss functions \cite{reb2,DSB,reb8}, designing complex regularization terms \cite{mu2023fedproc,huang3,lee2022preservation}, or aggregation strategies \cite{chen2023elastic,li2023revisiting,palihawadana2022fedsim}, primarily focus on post-hoc compensation at the optimization level. These methods attempt to indirectly offset the impact of data bias by adjusting the learning process, but they struggle to fundamentally bridge the gap between local observations and the ideal distribution \cite{fl3,GGEUR}. Another approach is data augmentation \cite{MetaSAug,zhou2023fedfa,H2T}, which expands the dataset by generating new samples. However, most augmentation methods (e.g., geometric transformations \cite{Gistnet}, Mixup \cite{mixup}) focus on the local neighborhood characteristics of samples or the combination of existing features, lacking the capability to model and recover the global structural properties of the entire class distribution \cite{GCL,FDC,GGEUR}. Consequently, when the observed distribution is not representative, these methods may generate spurious samples that deviate from the ideal distribution, potentially even degrading model performance \cite{OFA,GCL,FUR}.

\begin{figure*}[t]
\vskip -0.1in
\centering
\centerline{\includegraphics[width=1.85\columnwidth]{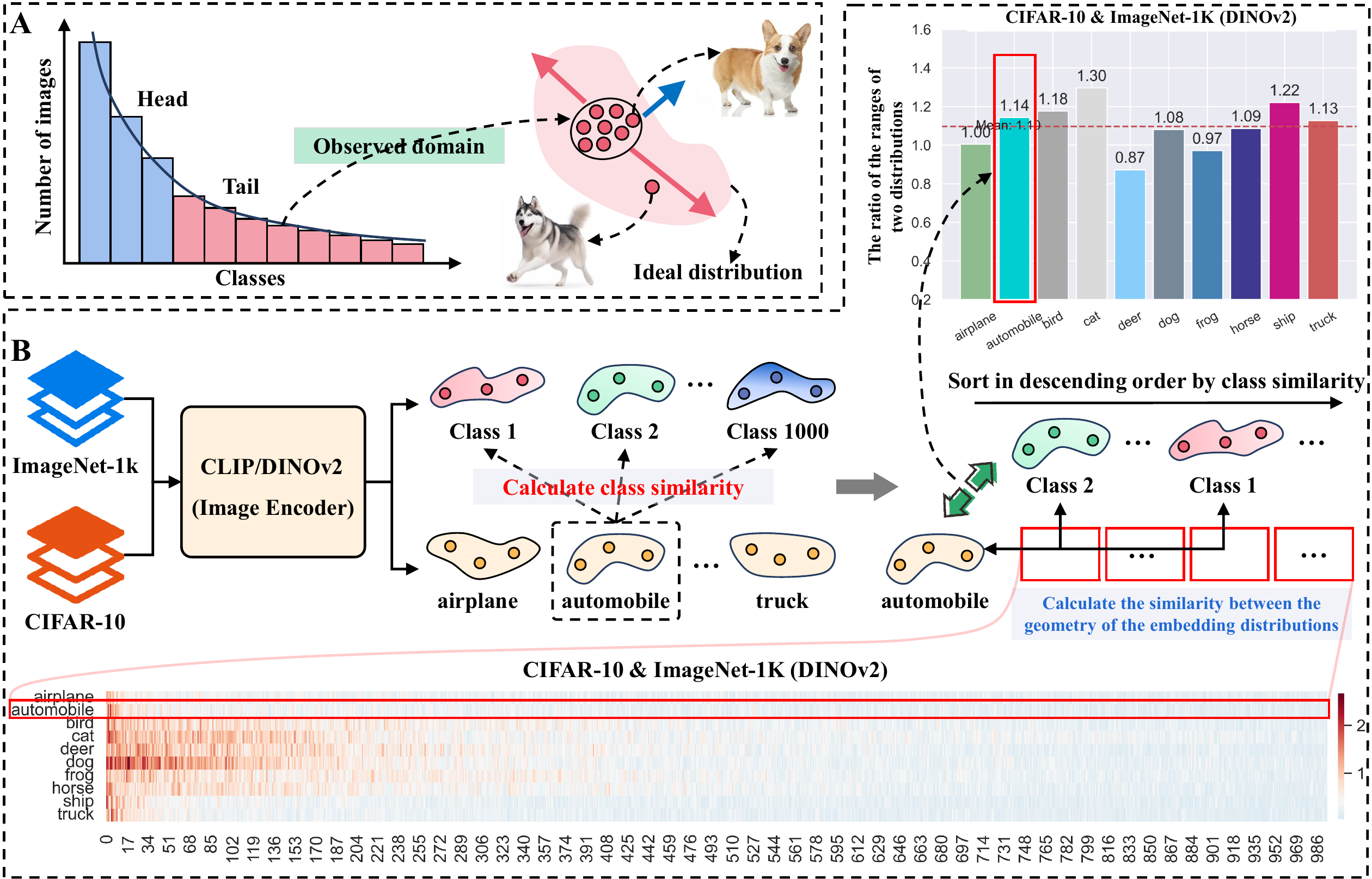}}
\vskip -0.05in
\caption{A. Samples at different positions within an embedding distribution possess distinct physical characteristics. B. Utilizing the foundation model to represent all classes across different datasets, we calculate the similarity between classes from different datasets and the similarity of geometric shapes and sizes between embedding distributions.}
\label{fig1}
\vskip -0.1in
\end{figure*}

Recovering the missing distribution from a limited number of available samples is a challenging task \cite{OFA,FDC}. As illustrated by the red distribution in Fig.~\ref{fig1}A, generating new samples along the red arrow direction is more likely to simulate the ideal distribution than along the blue arrow, given the existing observed distribution. Therefore, we identify determining a reasonable direction and scope for augmentation as the \textbf{key objective} for overcoming this challenge.
Recently, Vision Foundation Models (VFM), such as CLIP \cite{CLIP} and DINOv2 \cite{DINOv2}, have demonstrated remarkable capabilities in zero-shot recognition and cross-domain transfer tasks \cite{tpami7,Tip-adapter,CoOp}, suggesting that their embedding spaces contain deep semantic knowledge about visual concepts that transcends specific datasets. We propose a core insight: the semantics of a class are not only defined by its prototype but are also characterized by its geometric morphology in the embedding space. As shown in Fig.~\ref{fig1}A, samples at different positions within the same embedding distribution possess different physical characteristics. Through empirical studies, we discover that for semantically similar classes, even if they originate from completely different data domains, the distributions they form in the embedding space of foundation models (e.g., CLIP, DINOv2) exhibit highly similar geometric shapes (Fig.~\ref{fig1}B). We term this phenomenon \emph{Cross-Domain Geometric Consistency} (Section \ref{sec2.2}). This finding opens up a new possibility for addressing the distribution missing problem: we may leverage the foundation model as a geometric knowledge extractor, to extract the geometric prior of the ideal distribution and transfer it to data-scarce scenarios for distribution calibration.

Inspired by this, we propose a unified geometric knowledge-guided framework to address various distribution missing challenges (Section \ref{sec4}). In the federated learning scenario, privacy-preserving constraints prevent the centralization of all client samples to quantify the geometric shape of the global distribution. To this end, we propose a secure aggregation mechanism: each client computes its local covariance matrix and uploads it; the server then approximates the global covariance matrix by weighted aggregation of these local statistics, thereby constructing a geometric knowledge base that represents the ideal global distribution. This knowledge base is disseminated to the clients to guide the generation of virtual samples that conform to the global geometric shape, simulating a data distribution on the client that is closer to the ideal global distribution, and fundamentally alleviating the optimization conflicts caused by data heterogeneity (Section \ref{sec4.2}). In the long-tailed recognition scenario, for sample-scarce tail classes, the model can match semantically similar head classes from an external large-scale dataset (e.g., ImageNet) via the foundation model and transfer their complete geometric knowledge, thereby imagining and recovering the complete distribution that the tail classes should ideally possess (Section \ref{sec4.3}).

This paper shifts the focus of solving the distribution missing problem from optimization compensation and local enhancement to a new paradigm of knowledge transfer and distribution reconstruction. Our main contributions are summarized as follows\footnote{\textbf{This work extends our preliminary conference paper presented at CVPR (Oral) \cite{GGEUR} in the following aspects:}
(1) The significance of this study lies in elevating the focus of solving the distribution missing problem from the traditional levels of optimization compensation and local enhancement to a new paradigm of knowledge transfer and distribution reconstruction. We unify the data silos in federated learning and the sample scarcity in long-tailed learning as specific manifestations of distribution missing, thereby transforming the research motivation from a single-task driver into an exploration of a universal challenge.
(2) This paper presents a core insight with greater theoretical depth---cross-domain geometric consistency in the embedding space of vision foundation models---through extensive new experiments. This discovery serves as the theoretical cornerstone for the method's simultaneous application to both federated learning and long-tailed recognition, a deep mechanism not addressed in the original work.
(3) The original work primarily focused on distribution alignment within the federated learning. This study proposes a general knowledge transfer and distribution calibration paradigm, and successfully extends it to the domain of long-tailed recognition.
(4) This paper conducts experiments on a broader range of benchmarks, encompassing not only the federated learning datasets from the original paper but also adding multiple long-tailed recognition benchmarks.}:
\begin{itemize}
    \item We systematically discover and validate the existence of \emph{Cross-Domain Geometric Consistency} in the embedding space of vision foundation models, providing a solid empirical foundation for using foundation models for distribution calibration.
    \item We propose a geometric knowledge-guided distribution calibration method, treating distribution shape as a transferable prior, which unifies the solution to the distribution missing problem in both federated learning and long-tailed recognition.
    \item In the federated learning scenario, we propose a secure aggregation method based on local covariance matrices, enabling the efficient construction of global geometric knowledge while strictly preserving client data privacy.
    \item In the long-tailed recognition scenario, to address the issue of generated samples interrupting the training process, we innovatively propose a geometric knowledge-guided embedding uncertainty representation layer (\textbf{GGEUR-Layer}), achieving end-to-end joint optimization of distribution calibration and classifier training.
    \item We conducted extensive experiments on multiple challenging benchmark datasets (Section\ref{sec5}), demonstrating the effectiveness and generality of geometric knowledge-guided distribution calibration.
\end{itemize}

\section{Related Work}
\label{sec:related_work}

This section reviews existing research from the unified perspective of learning under data-constrained scenarios.

\subsection{Training-Stage Adaption}
\label{sec:opt_compensation}

When models face data-constrained challenges (e.g., sample scarcity, distribution skew), a direct approach is to compensate at the optimization level. These methods do not alter the data itself but adjust the learning process to mitigate the impact of data bias.

 \textit{Loss Re-weighting and Re-sampling}: In long-tailed learning, methods like Focal Loss \cite{reb1} and Class-Balanced Loss \cite{reb2} assign higher loss weights to tail classes to balance optimization \cite{reb8,yan2,DSB,CR,IGAM}. In federated learning, FedProx \cite{li2} and SCAFFOLD \cite{SCAFFOLD} introduce regularization terms into the local objective to constrain local updates from deviating too far from the global model \cite{moon,huang3,fedlc,feddecorr}, alleviating data heterogeneity. These methods are simple and effective but are essentially a post-hoc compensation for imbalanced data, failing to address the fundamental issue of the model's incomplete cognition of the ideal distribution.

 \textit{Decoupled Training}: Methods like Decoupling \cite{reb12} and BBN \cite{bbn} propose to decouple representation learning from classifier learning. They first learn a general representation on imbalanced data, then train the classifier on a re-sampled or re-weighted balanced set \cite{FDC,CMO,FUR}. This partially decouples the sensitivity of representation learning to data distribution, but the representation learning phase is still constrained by the incomplete observed data. Recent works like RIDE \cite{RIDE} and ResLT \cite{reslt} further improve tail-class performance by training multiple expert classifiers or leveraging self-supervised learning.
    
 \textit{Aggregation Strategy Optimization}: In federated learning, FedNova \cite{fednova} and FedOpt \cite{fedopt} normalize the magnitude or direction of local updates to mitigate aggregation bias caused by heterogeneity \cite{al2020federated,chen2023elastic,huang4}. These methods perform finer adjustments at the optimization level but still fail to address the core issue of data distribution mismatch.

\subsection{Data Augmentation: Expansion at the Data Level}
\label{sec:data_augmentation}

Another class of methods aims to directly expand the dataset to compensate for insufficient observations. However, these methods often lack explicit guidance on the ideal global distribution structure when generating new data.

 \textit{Transformation-based Augmentation}: Methods like Mixup \cite{mixup} and CutMix \cite{cutmix} generate new training samples by linearly interpolating or stitching existing samples. While they increase data diversity, their generation process is limited to linear combinations of existing samples, unable to create novel patterns beyond the current observation. UniMix \cite{unimix} and OFA \cite{OFA} extend Mixup to feature space, but still relies on linear interpolation.
    
 \textit{Generation-based Augmentation}: Leveraging generative models like GANs or Diffusion Models to generate new samples. While more powerful, issues like training instability and mode collapse make them difficult to apply in data-constrained scenarios. More importantly, these models themselves require large amounts of data for training, and the generation process lacks explicit constraints on the 'ideal distribution' shape, easily producing spurious samples that do not align with the true distribution. In federated learning, FedGen \cite{zhu2021data} and FedGAN \cite{fedgan} attempt to train a generator on the server, but this introduces additional communication overhead and privacy risks.

\subsection{External Knowledge Transfer}
\label{sec:knowledge_transfer}

Knowledge transfer aims to leverage external information to compensate for insufficient local data, representing one of the most promising directions for solving data-constrained problems. Existing work primarily focuses on two levels:

 \textit{Semantic Knowledge Transfer}: Leveraging the strong representational capabilities of pre-trained models (e.g., ImageNet pre-trained ResNet) or vision foundation models through fine-tuning or prompt learning \cite{CoOp,Tip-adapter,lpt} to transfer general knowledge to the target task. These methods enhance the initial representation quality but do not directly address the mismatch at the distribution level.
    
 \textit{Statistical Knowledge Transfer}: Some works attempt to transfer statistical information of the data \cite{OFA,CMO,FDC,FUR,GCL,H2T}. For instance, in long-tailed learning, FTL \cite{FTL} and FDC \cite{FDC} assume that the feature distributions of the common and UR classes (i.e., rare classes) have the same variance, so the variance from the head classes is used to guide the feature enhancement of the tail classes.  H2T \cite{H2T} proposes using head class features to construct tail class features. OFA \cite{OFA} decomposes the features of each class into class-generic features and class-specific features. During training, the tail class-specific features are fused with the head class-generic features to generate new features to augment the tail classes.

 \textit{Our Position}: Unlike the aforementioned works, we propose a novel knowledge transfer paradigm—geometric knowledge transfer. We do not transfer model parameters, raw samples, or simple statistics, but rather the geometric shape of the class distribution—a deep prior knowledge encoded by vision foundation models. \textbf{Compared to} the Gaussian assumption, the geometric shapes provide a more accurate description of embedding distributions. This knowledge is cross-domain robust and can provide powerful, structured guidance for calibrating distributions in data-constrained scenarios, thereby elevating knowledge transfer from the representation level and statistical level to the distribution structure level.


\section{Foundation Model as Bridges for Transferring Geometric Knowledge}
\label{sec2}

To effectively leverage the geometric shape of a distribution to assist in recovering the ideal distribution of rare classes, we must carefully consider three core questions: 
\begin{itemize}
\item[(1)] \textbf{Similarity}: Is the geometric shape being transferred highly similar to the ideal distribution's geometric shape of the target category? 
\item[(2)] \textbf{Size Consistency}: Is the size (Scale) of the transferred distribution close to that of the target's ideal distribution? Because a discrepancy in size will directly impact whether the reconstructed distribution can adequately cover its true counterpart. 
\item[(3)] \textbf{Matching Stability}: For the same class, when its samples transition from being scarce to sufficient, does the most similar external class it matches remain consistent? This relates to the reliability of the transferred knowledge.
\end{itemize}

To systematically investigate the above questions, we employ the following experimental design: For the first two questions (Section \ref{sec2.2}), we use the complete, balanced versions of CIFAR-10-LT and CIFAR-100-LT (i.e., CIFAR-10 and CIFAR-100) to obtain the ideal distributions of the rare classes as a baseline. For the third question (Section \ref{sec2.3}), we will conduct a comparative analysis of the consistency of category matching results between the long-tailed datasets (CIFAR-10-LT/CIFAR-100-LT) and their complete versions.

In this section, we first define the geometric shape, size, and similarity measures of embedding distributions. Subsequently, through empirical studies on multiple benchmark datasets, we reveal a key phenomenon: Vision Foundation Models (VFM) can represent semantically similar categories from different data domains as embedding distributions with similar geometric shapes and sizes. We term this phenomenon \textit{Cross-Domain Geometric Consistency}. This discovery constitutes the cornerstone of our \textit{geometric knowledge-guided distribution reconstruction} paradigm, providing a solid empirical foundation for solving the \textit{distribution missing} problem through external knowledge transfer.

\begin{figure*}[t]
\centering
\centerline{\includegraphics[width=2\columnwidth]{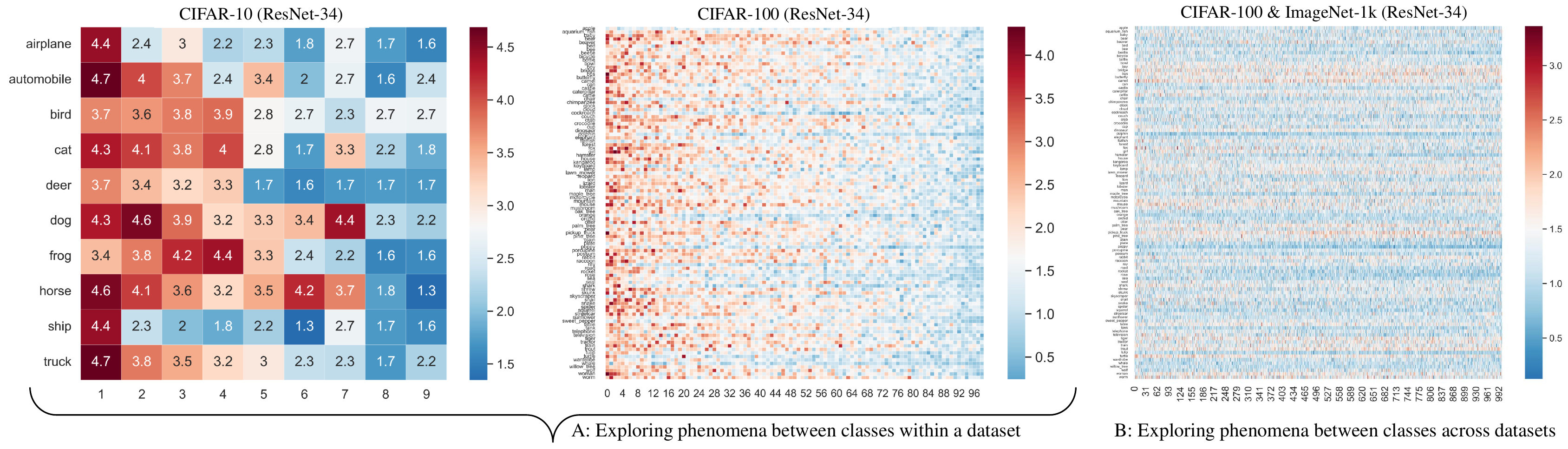}}
\vskip -0.1in
\caption{The horizontal coordinate is the index of classes, indicating from left to right the classes that are most similar to the class in the vertical coordinate to the least similar, respectively. Each element represents the similarity between the geometric shapes of the embedding distributions.}
\label{fig2}
\end{figure*}

\subsection{Geometric Shape, Size, and Similarity of Embedding Distributions}
\label{sec2.1}

To formally characterize the structural properties of data distributions in the embedding space, we define their geometric shape and size based on the eigendecomposition of the covariance matrix.

\begin{definition}[\textbf{Geometric Shape}]
Let $X = [\mathbf{x}_1, ..., \mathbf{x}_n] \in \mathbb{R}^{P \times n}$ be a set of $n$ samples from a specific class in a $P$-dimensional embedding space. The covariance matrix $\Sigma_X$ of this distribution is estimated as:
\begin{equation}
\Sigma_X=\mathbb{E}\left[\frac{1}{n} {\textstyle \sum_{i=1}^{n}x_ix_i^T}\right ]=\frac{1}{n}XX^T\in \mathbb{R}^{P\times P}.
\end{equation}
Performing eigendecomposition on $\Sigma_X$ yields $P$ non-negative eigenvalues $\lambda_1 \geq \lambda_2 \geq ... \geq \lambda_P$ and their corresponding orthogonal eigenvectors $\boldsymbol{\xi}_1, \boldsymbol{\xi}_2, ..., \boldsymbol{\xi}_P$. These eigenvectors define the principal axes of the data distribution. Therefore, we define the \textbf{geometric shape} $GD_X$ of data $X$ as the set of the first $m$ eigenvectors corresponding to the $m$ largest eigenvalues:
\begin{equation}
GD_X = \{\boldsymbol{\xi}_1, \boldsymbol{\xi}_2, ..., \boldsymbol{\xi}_m\}.
\end{equation}
In this work, we set $m=5$ by default.
\end{definition}

\begin{definition}[\textbf{Size}]
The eigenvalue $\lambda_i$ represents the variance of the data along the direction of its corresponding eigenvector $\boldsymbol{\xi}_i$. The overall \textbf{size} $S(X)$ of the distribution is thus measured by the sum of all eigenvalues, which is the trace of the covariance matrix:
\begin{equation}
S(X) = \sum_{i=1}^{P} \lambda_i = \mathrm{tr}(\Sigma_X).
\end{equation}
$S(X)$ reflects the overall "scale" or "spread" of the embedding distribution.
\end{definition}

\begin{definition}[\textbf{Geometric Shape Similarity}]
Given two distributions $X_1$ and $X_2$ with geometric shapes $GD_{X_1} = \{\boldsymbol{\xi}_{X_1}^1, ..., \boldsymbol{\xi}_{X_1}^m\}$ and $GD_{X_2} = \{\boldsymbol{\xi}_{X_2}^1, ..., \boldsymbol{\xi}_{X_2}^m\}$, their geometric shape similarity $Sim(GD_{X_1}, GD_{X_2})$ is defined as the normalized sum of the absolute values of the dot products between corresponding eigenvectors:
\begin{equation}
Sim(GD_{X_1}, GD_{X_2}) = \sum_{i=1}^{m} |\langle \boldsymbol{\xi}_{X_1}^i, \boldsymbol{\xi}_{X_2}^i \rangle|.
\end{equation}
This similarity metric ranges from $0$ to $m$, with a value closer to $m$ indicating a higher degree of geometric shape alignment.
\end{definition}

In the embedding space induced by VFM, the principal directions of class distributions are not only ordered by variance but also tend to exhibit semantic hierarchy and cross-class alignment. Prior studies \cite{CLIP,dalle2} have shown that the first few principal directions often capture high-level semantic structures, such as global shape and texture patterns, while subsequent components encode finer-grained or less dominant variations. This structured representation suggests that, for semantically similar classes, the ordering and semantic meaning of principal directions may remain consistent: the first principal direction corresponds to the most prominent semantic variation, the second to secondary structures, and so on. Therefore, aligning principal directions by index for similarity measurement is not merely a computational simplification but also possesses inherent semantic plausibility.

\subsection{Cross-Domain Geometric Consistency}
\label{sec2.2}

\subsubsection{Small Models Cannot Associate Geometric Shapes Across Data Domains}
\label{sec2.2.1}

We train a standard ResNet-34 \cite{ResNet} on CIFAR-10 \cite{CIFAR} and extract $p$-dimensional image embeddings from its last hidden layer for each class. Let $Z_i = [z_i^1, \dots, z_i^n] \in \mathbb{R}^{p \times n}$ be the embedding set for class $i$, with its distribution center $C_i = \frac{1}{n} \sum_{k=1}^{n} z_i^k \in \mathbb{R}^{p \times 1}$. Given the embedding set $Z_j$ for class $j$, the similarity between class $i$ and $j$ is computed as $\frac{C_i^\top C_j}{\|C_i\| \cdot \|C_j\|}$, where a higher value indicates greater similarity. We first calculate the similarity between each class and all others, then sort the classes in descending order of their similarity to a reference class. Next, we compute the similarity of the geometric shapes (as defined in Section~\ref{sec2.1}) between the reference class and each other class in this sorted order. The results, plotted in Fig.~\ref{fig2}A, show that as the class similarity increases, the similarity of their embedding distribution geometries also tends to increase. We verify this phenomenon on the CIFAR-100 dataset with the same experimental setup.

However, we find that this phenomenon does not hold when matching similar classes across different datasets. We train ResNet-34 models on CIFAR-100 and ImageNet-1k \cite{imagenet} separately and extract their image embeddings. We then calculate the class similarity between each class in CIFAR-100 and all classes in ImageNet, sorting the ImageNet classes in descending order of similarity. According to this order, we compute the geometric shape similarity between the CIFAR-100 class and each ImageNet class. The experimental results in Fig.~\ref{fig2}B do not exhibit the same increasing trend observed in Fig.~\ref{fig2}A. These results collectively confirm that conventional small models cannot associate geometric knowledge across different datasets.

\begin{figure}[t]
\centering
\centerline{\includegraphics[width=1\columnwidth]{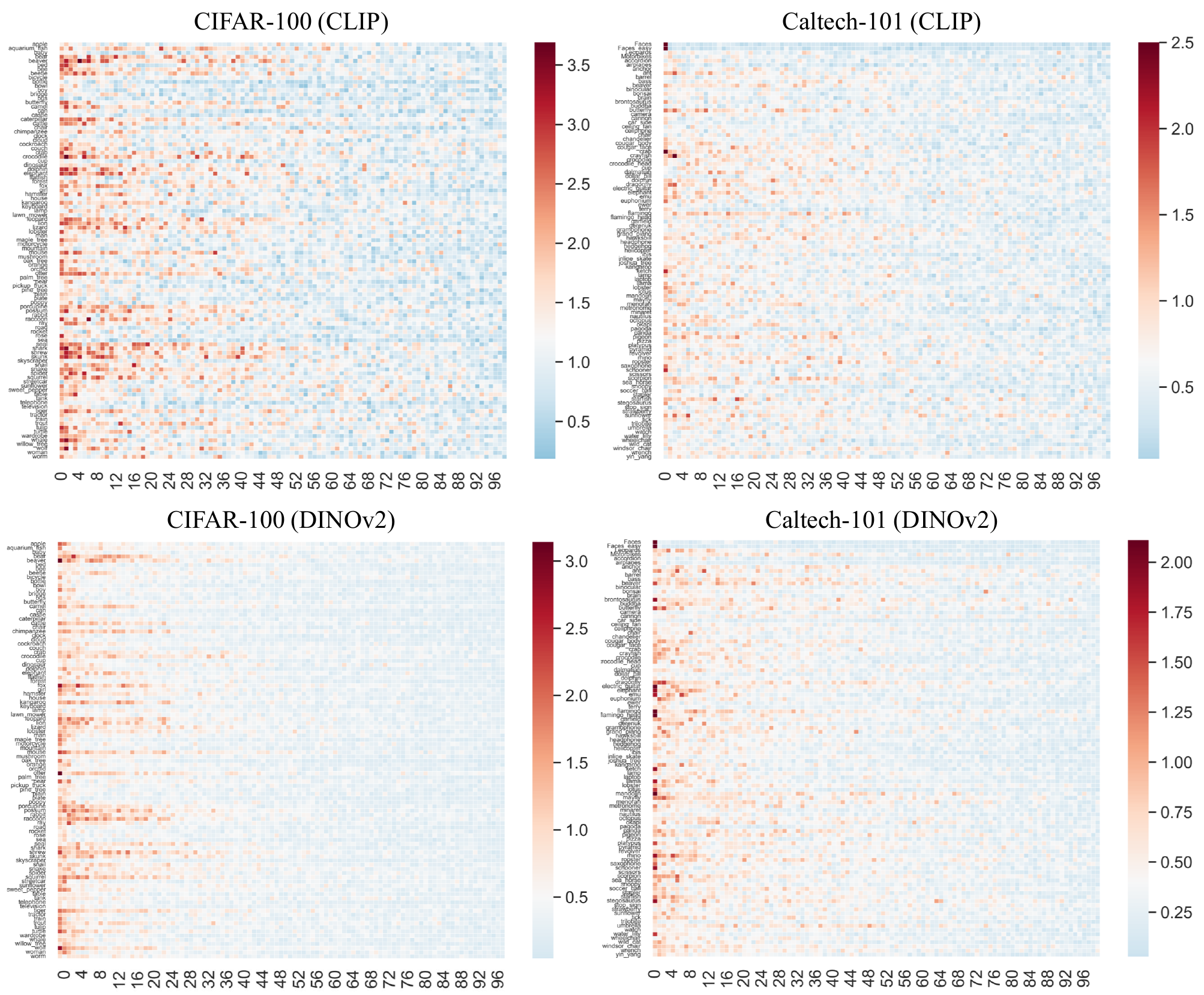}}
\vskip -0.1in
\caption{Calculate the similarity between classes within the dataset and the similarity of geometric shapes between corresponding embedding distributions.}
\label{fig3}
\vskip -0.1in
\end{figure}

\begin{figure}[t]
\centering
\centerline{\includegraphics[width=1\columnwidth]{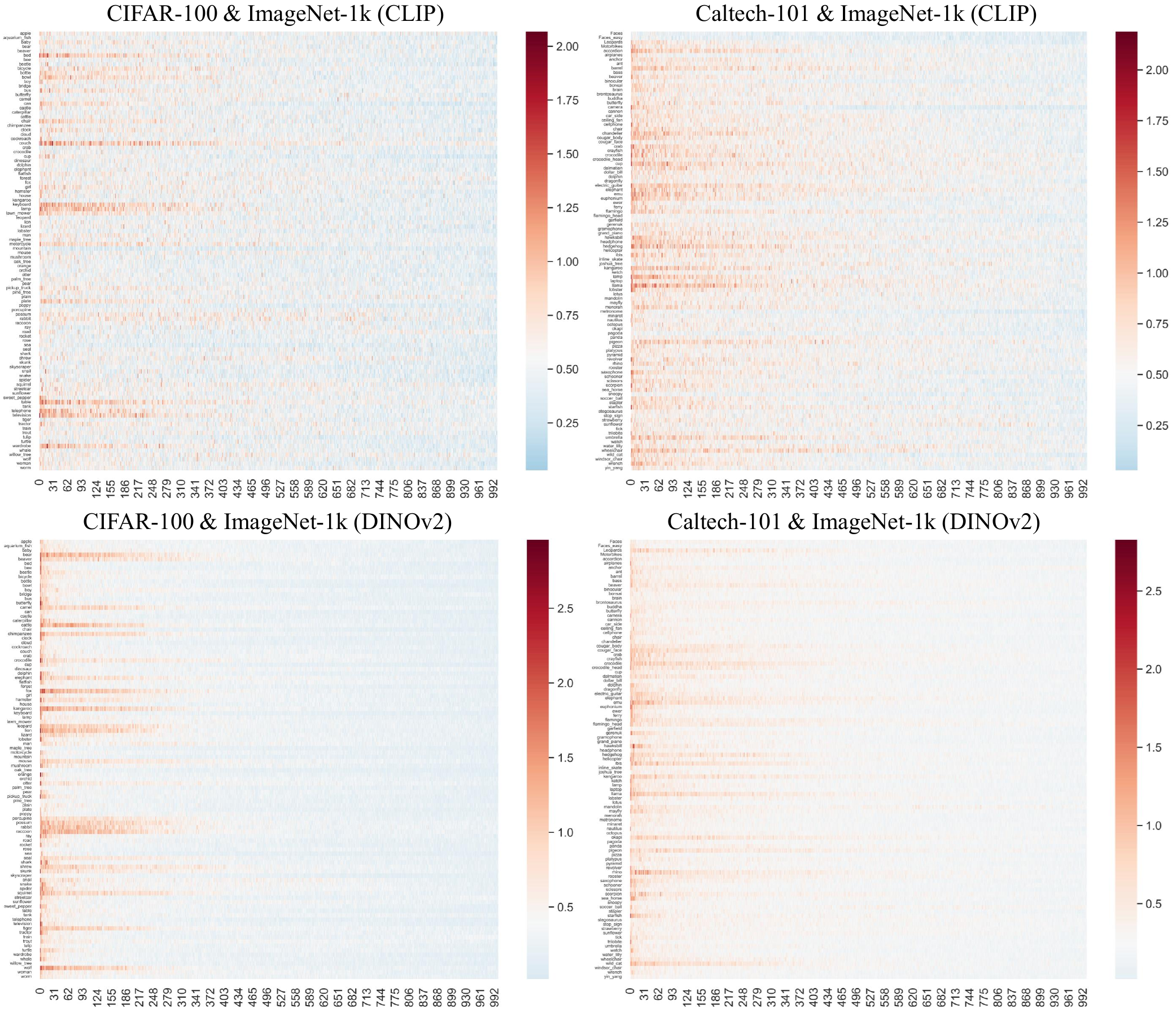}}
\vskip -0.1in
\caption{Compute the similarity between classes belonging to different datasets and the similarity of geometric shapes between corresponding embedding distributions.}
\label{fig4}
\vskip -0.1in
\end{figure}

\begin{figure*}[h]
\centering
\centerline{\includegraphics[width=2\columnwidth]{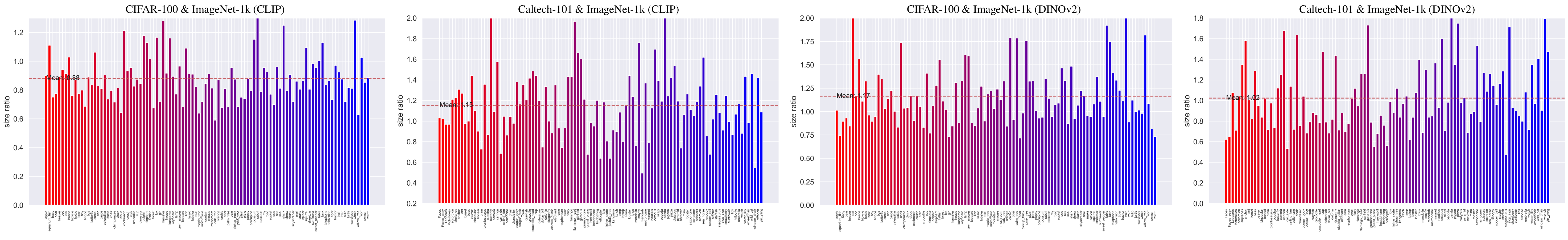}}
\vskip -0.1in
\caption{For two classes from different datasets with high similarity, the ratio of their embedding distributions sizes approaches $1$.}
\label{fig5}
\vskip -0.1in
\end{figure*}

\begin{figure*}[t]
\centering
\centerline{\includegraphics[width=2\columnwidth]{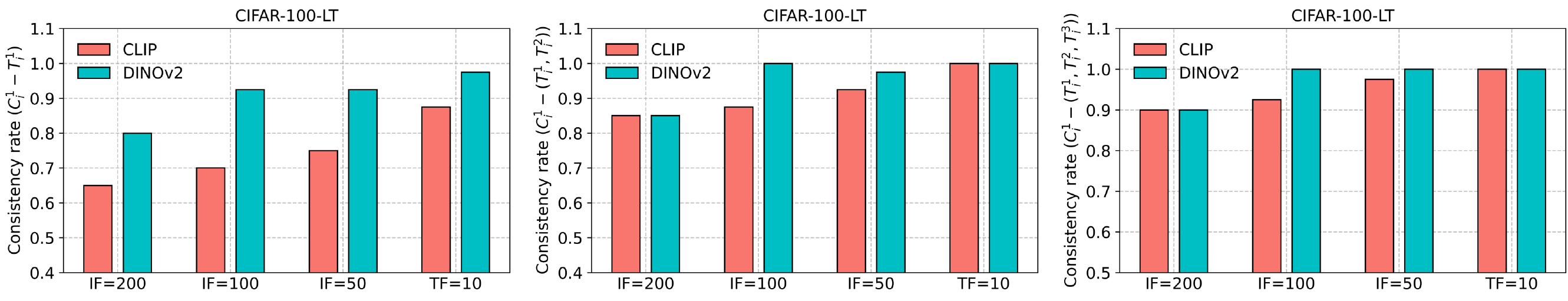}}
\vskip -0.05in
\caption{Match the most similar classes in ImageNet-1k for the $40$ tail classes of CIFAR-100-LT as well as their full versions, and calculate the proportion of tail classes that agree with the most similar class matched to their full versions.}
\label{fig6}
\end{figure*}

\subsubsection{Vision Foundation Models: Bridging Geometric Knowledge Across Data Domains}
\label{sec2.2.2}

The preceding experiments have demonstrated that small models cannot associate geometric knowledge across data domains. Given the powerful representational capabilities of Vision Foundation Models (VFM), we aim to leverage them for cross-dataset geometric knowledge transfer. Below, \textbf{we primarily focus on two aspects:}
\begin{itemize}
\item[(1)] Within a single dataset, do VFM exhibit the same phenomenon as small models, where more similar classes have more similar geometric shapes?
\item[(2)] Can VFM demonstrate a phenomenon where, for two classes from different datasets, the more similar they are, the more similar their embedding distribution geometric shapes tend to be?
\end{itemize}

We selected CLIP \cite{CLIP} and DINOv2 (ViT-B/16) \cite{DINOv2} as the two Vision Foundation Models to investigate the transferability of geometric shapes. Firstly, we extracted image embeddings from CIFAR-100 \cite{CIFAR} and Caltech-101 \cite{Clatech101} using CLIP and DINOv2 separately. Subsequently, we conducted experiments similar to those in Fig.~\ref{fig2}, wherein we matched similar classes within each dataset and computed the similarity of geometric shapes between the embedding distributions of corresponding classes. The experimental results are depicted in Fig.~\ref{fig3}, showing that Vision Foundation Models also exhibit the phenomenon of geometric shapes of embedding distributions becoming more similar as classes become more similar within datasets. It is noteworthy that compared to CLIP, DINOv2 shows a more pronounced phenomenon. This may be attributed to DINOv2 being trained through multi-level masked self-supervised learning from pure visual data. It pays attention to finer details of visual information, enabling a more refined representation of the geometric shapes of embedding distributions for each class, thus making the phenomenon more evident.

Further, we selected ImageNet-1k \cite{imagenet} as an external knowledge base. We calculated the semantic similarity between each class in CIFAR-100 and Caltech-101 and all classes in ImageNet-1k, as well as the geometric shape similarity between their corresponding embedding distributions. Following the same procedure as in Fig.~\ref{fig3}, we plotted the results in Fig.~\ref{fig4}. To our surprise, we found that as the class similarity increases, the similarity of the geometric shapes between the corresponding embedding distributions also tends to increase. This suggests that Vision Foundation Models can serve as a bridge for associating geometric knowledge across different datasets. Particularly, the phenomenon observed with DINOv2 is more pronounced, leading us to speculate that the geometric shapes represented by DINOv2 could more effectively and accurately guide the recovery of the ideal distribution of rare classes.

When assisting rare classes, it is also important to consider the size of the distribution being transferred. We matched each class from CIFAR-10, CIFAR-100, and Caltech-101 with its most similar class from ImageNet-1k, and then computed the ratio of the sizes ($S(\cdot)$, as defined in Section~\ref{sec2.1}) between their corresponding embedding distributions. The experimental results, as shown in Fig.~\ref{fig5}, reveal that the distributions represented by both CLIP and DINOv2 exhibit similar sizes for the embedding distributions of highly similar classes. These findings collectively provide a strong foundation for our proposed method of rare classes distribution recovery.

\subsection{Matching Consistency of Similar Classes with Sufficient vs. Insufficient Samples}
\label{sec2.3}

The study above was conducted under the assumption of sufficient samples, focusing on the geometric similarity between ideal true distributions. Suppose we wish to recover the ideal distribution of rare classes in CIFAR-100-LT; an ideal approach would be to match each rare class with its most similar class in ImageNet-1k and then transfer the geometric shape and size of the embedding distribution from the matched ImageNet class to the rare class (the transfer mechanism will be detailed in Section~\ref{sec4}). However, a crucial prerequisite for the feasibility of this approach is that the highly similar class matched for a rare class in ImageNet-1k remains consistent with the class matched when the rare class has sufficient samples.

We continue to use CLIP \cite{CLIP} and DINOv2 \cite{DINOv2} to extract the embedding distributions of the classes. We match the most similar classes $C_1^1,\dots,C_{40}^1$ in ImageNet-1k for the $40$ rare classes $C_1,\dots,C_{40}$ (with the fewest samples) in CIFAR-100-LT, respectively. Concurrently, we identify the complete versions $T_1,\dots,T_{40}$ of these $40$ rare classes in the balanced CIFAR-100 dataset, and then match them with their first, second, and third most similar classes $T_i^1,T_i^2,T_i^3$ ($i=1,\dots,40$) in ImageNet-1k. We check whether $C_i^1$ matches $T_i^1$ for $i=1,\dots,40$, and calculate the proportion of exact matches out of the $40$ classes. The experimental results are plotted in Fig.~\ref{fig6}. The ideal scenario is a $100\%$ consistency rate between the sets $\{C_1^1,\dots,C_{40}^1\}$ and $\{T_1^1,\dots,T_{40}^1\}$.

However, this is a stringent requirement. For a given class, the geometric shapes of its second and third most similar classes are also highly similar to its own. Therefore, we relax the criterion. We calculate the proportion of cases where $C_i^1$ is contained within the top-2 matches $\{T_i^1, T_i^2\}$, and the proportion where $C_i^1$ is contained within the top-3 matches $\{T_i^1, T_i^2, T_i^3\}$ for $i=1,\dots,40$. The results in Fig.~\ref{fig6} show that the most similar classes matched for rare classes from the external dataset (ImageNet-1k) are highly consistent with the top similar classes matched when the rare classes have sufficient samples. This ensures the feasibility and reliability of our method for finding and transferring geometric knowledge based on class similarity.

\section{Geometric Knowledge-Guided Distribution Calibration}
\label{sec4}

We have empirically demonstrated the existence of cross-domain geometric consistency in the embedding space of Vision Foundation Models. Building upon this foundation, this section proposes a unified framework for geometric knowledge-guided distribution calibration. The framework aims to fundamentally calibrate the observed distributions in data-constrained scenarios by securely constructing and transferring geometric knowledge, thereby enabling them to approximate the ideal distribution.

\subsection{Framework Overview}
\label{sec4.1}

Our framework consists of two core stages: 
\begin{itemize}
\item[(1)] \textit{Construction of the Global Geometric Knowledge Base}.
\item[(2)] \textit{Geometric Knowledge-Guided Distribution Calibration}. 
\end{itemize}
The generality of this framework lies in its core idea---"transferring geometric knowledge as a prior"---which can be seamlessly applied to diverse scenarios such as federated learning and long-tailed recognition.

In \textbf{federated learning} (Section \ref{sec4.2}), the first stage occurs at the server. The server constructs a geometric knowledge base representing the ideal global distribution by securely aggregating local geometric statistics from all clients. In the second stage, this knowledge base is disseminated to the clients to guide local distribution alignment. In \textbf{long-tailed recognition} (Section \ref{sec4.3}), the knowledge base for the first stage is directly served by an external large-scale dataset (e.g., ImageNet). In the second stage, through cross-domain matching, the geometric knowledge of similar classes from the external knowledge base is transferred to the rare classes to simulate their ideal distribution.

\subsection{Local Distribution Calibration in Federated Data Heterogeneity Scenarios}
\label{sec4.2}

In this section, we address a critical issue: how to approximate the geometric shape of the global data distribution using local client data while preserving privacy (Section \ref{sec4.2.1}). We then detail how to simulate the ideal global distribution locally in both single-domain (Section \ref{sec4.2.2}) and multi-domain scenarios (Section \ref{sec4.2.2}) by leveraging the geometric shape of the global distribution.

\subsubsection{Computation of Global Geometric Shapes under Privacy Constraints}
\label{sec4.2.1}

In federated learning, privacy-preserving constraints prohibit the centralization of clients' raw data, making it impossible to directly compute the geometric shape of the global distribution. To address this challenge, we design a secure and efficient aggregation mechanism to approximate the global geometric shape without compromising data privacy.

Consider a classification task with $K$ clients and $C$ classes. Let $n_k^i$ denote the number of samples belonging to class $i$ on client $k$, with the sample set represented as $\{ x_k^{i,1}, \dots, x_k^{i,n_k^i} \}$. The global distribution for class $i$ is formed by aggregating the local samples from all $K$ clients. As established in Section~\ref{sec2.1}, the geometric shape of a distribution is derived from its covariance matrix. Our goal is therefore to approximate the global covariance matrix $\Sigma_i$ for each class $i$ using only locally computed statistics.

We propose to estimate $\Sigma_i$ by leveraging the local covariance matrices and means from all clients. For class $i$, each client $k$ first computes its local mean $\mu_k^i$ and local covariance matrix $\Sigma_k^i (k=1,\dots,K)$:
\begin{equation}
\setlength\abovedisplayskip{2pt} 
\setlength\belowdisplayskip{2pt}
\mu_k^i = \frac{1}{n_k^i} \sum_{j=1}^{n_k^i} x_k^{i,j}, \quad
\Sigma_k^i = \frac{1}{n_k^i} \sum_{j=1}^{n_k^i} (x_k^{i,j} - \mu_k^i)(x_k^{i,j} - \mu_k^i)^\top.
\label{eq:local_stats}
\end{equation}
The global covariance matrix $\Sigma_i$ is the covariance of the combined data from all clients, defined as:
\begin{equation}
\setlength\abovedisplayskip{2pt} 
\setlength\belowdisplayskip{2pt}
\begin{split}
\mu_i = \frac{1}{N_i} \sum_{k=1}^K \sum_{j=1}^{n_k^i} x_k^{i,j} = \frac{1}{N_i} \sum_{k=1}^K n_k^i \mu_k^i, \\
\Sigma_i = \frac{1}{N_i} \sum_{k=1}^K \sum_{j=1}^{n_k^i} (x_k^{i,j} - \mu_i)(x_k^{i,j} - \mu_i)^T,
\label{eq:global_def}
\end{split}
\end{equation}
where \( N_i = \sum_{k=1}^K n_k^i \) represents the total number of samples in class \( i \) across all clients. We can decompose \( (x_k^{i,j} - \mu_i) \) as \( (x_k^{i,j} - \mu_k^i) + (\mu_k^i - \mu_i) \), so the global covariance matrix \( \Sigma_i \) can be rewritten as:
\begin{small}
\begin{equation}
\setlength\abovedisplayskip{2pt} 
\setlength\belowdisplayskip{2pt}
\begin{split}
\Sigma_i = \frac{1}{N_i} \sum_{k=1}^K \sum_{j=1}^{n_k^i} [ (x_k^{i,j} - \mu_k^i + \mu_k^i - \mu_i)(x_k^{i,j} - \mu_k^i + \mu_k^i - \mu_i)^T ].
\end{split}
\end{equation}
\end{small}
Expanding the above equation yields:
\begin{small}
\begin{equation}
\setlength\abovedisplayskip{2pt} 
\setlength\belowdisplayskip{2pt}
\begin{split}
\Sigma_i = \frac{1}{N_i} \sum_{k=1}^K \sum_{j=1}^{n_k^i} [ (x_k^{i,j} - \mu_k^i)(x_k^{i,j} - \mu_k^i)^T + (x_k^{i,j} - \mu_k^i)(\mu_k^i - \mu_i)^T  \\
+ (\mu_k^i - \mu_i)(x_k^{i,j} - \mu_k^i)^T + (\mu_k^i - \mu_i)(\mu_k^i - \mu_i)^T].
\end{split}
\end{equation}
\end{small}
By the properties of covariance matrices, the first term is the local covariance matrix \( \Sigma_k^i \), and the expected values of the second and third terms are zero. The fourth term can be computed as:
\begin{small}
\begin{equation}
\setlength\abovedisplayskip{2pt} 
\setlength\belowdisplayskip{2pt}
\begin{split}
\sum_{j=1}^{n_k^i} (\mu_k^i - \mu_i)(\mu_k^i - \mu_i)^T = n_k^i (\mu_k^i - \mu_i)(\mu_k^i - \mu_i)^T.
\end{split}
\end{equation}
\end{small}
Thus, the global covariance matrix can be obtained by combining the local covariance matrices and local means as:
\begin{small}
\begin{equation}
\setlength\abovedisplayskip{3pt} 
\setlength\belowdisplayskip{3pt}
\begin{split}
\Sigma_i = \frac{1}{N_i} \left( \sum_{k=1}^K n_k^i \Sigma_k^i + \sum_{k=1}^K n_k^i (\mu_k^i - \mu_i)(\mu_k^i - \mu_i)^T \right).
\end{split}
\end{equation}
\end{small}
This formulation allows the server to compute the global covariance matrix $\Sigma_i$ by collecting only the local means $\mu_k^i$ and local covariance matrices $\Sigma_k^i$ from the clients. The geometric shape of the global distribution for class $i$ is then obtained via eigendecomposition of $\Sigma_i$, all without any client sharing its raw data.

We now discuss the privacy guarantees of this aggregation mechanism. The server disseminates only the eigenvectors and eigenvalues of $\Sigma_i$ back to the clients, which is not enough information to reconstruct any client's original data for the following reasons:

\textbf{(1) Eigenvectors and Eigenvalues Do Not Contain Raw Data.}
The eigendecomposition reveals only the geometric structure (principal axes and variances) of the data distribution, not the individual data points. Reconstructing the original dataset from its covariance matrix is an ill-posed problem with infinitely many solutions.

\textbf{(2) Low-Rank Property Prevents Data Reconstruction.}
The covariance matrix $\Sigma_i$ is typically \textbf{low-rank} ($\text{rank}(\Sigma_i) \ll d$, where $d$ is the embedding dimension). This means that even with full knowledge of the eigenvectors and eigenvalues, one can only recover the data's projection onto a low-dimensional subspace, not its full details.

\textbf{(3) Aggregation Prevents Isolation of Individual Client Contributions.}
The global covariance matrix is a weighted sum of contributions from all clients. Because these contributions are mixed through aggregation:
\begin{itemize}
    \item No single client can isolate another client’s contribution from the shared geometric knowledge.
    \item The removal of a single client's data affects the global $\Sigma_i$ in a distributed manner across all eigenvectors, making the individual influence indistinguishable.
\end{itemize}

\textbf{(4) Existing Literature Supports the Privacy of Covariance Matrices.}
Prior works have confirmed that sharing higher-order statistics like covariance matrices poses significantly lower privacy risks compared to sharing model gradients \cite{zhou2023fedfa}.

We describe below how geometric knowledge can be used to simulate the ideal global data distribution locally in both single-domain and multi-domain scenarios.

\subsubsection{Single-Domain Federated Learning}
\label{sec4.2.2}

When the data originates from a single domain, we primarily address the issue of label skew. Before federated learning begins, each client uses CLIP to extract \( p \)-dimensional embeddings of its local data. Each client then locally computes the per-class local covariance matrices and local sample means, which are subsequently uploaded to the server. For each class, the server generates a global covariance matrix, \( \Sigma_1, \Sigma_2, \ldots, \Sigma_C \), using Equation (\textcolor{red}{10}). Next, each class’s global geometric shape is quantified based on Definition (\textcolor{red}{1}), and these global geometric shapes (i.e., eigenvalues and eigenvectors) are sent to each client.

\begin{figure}[t]
  \centering
   \includegraphics[width=1\linewidth]{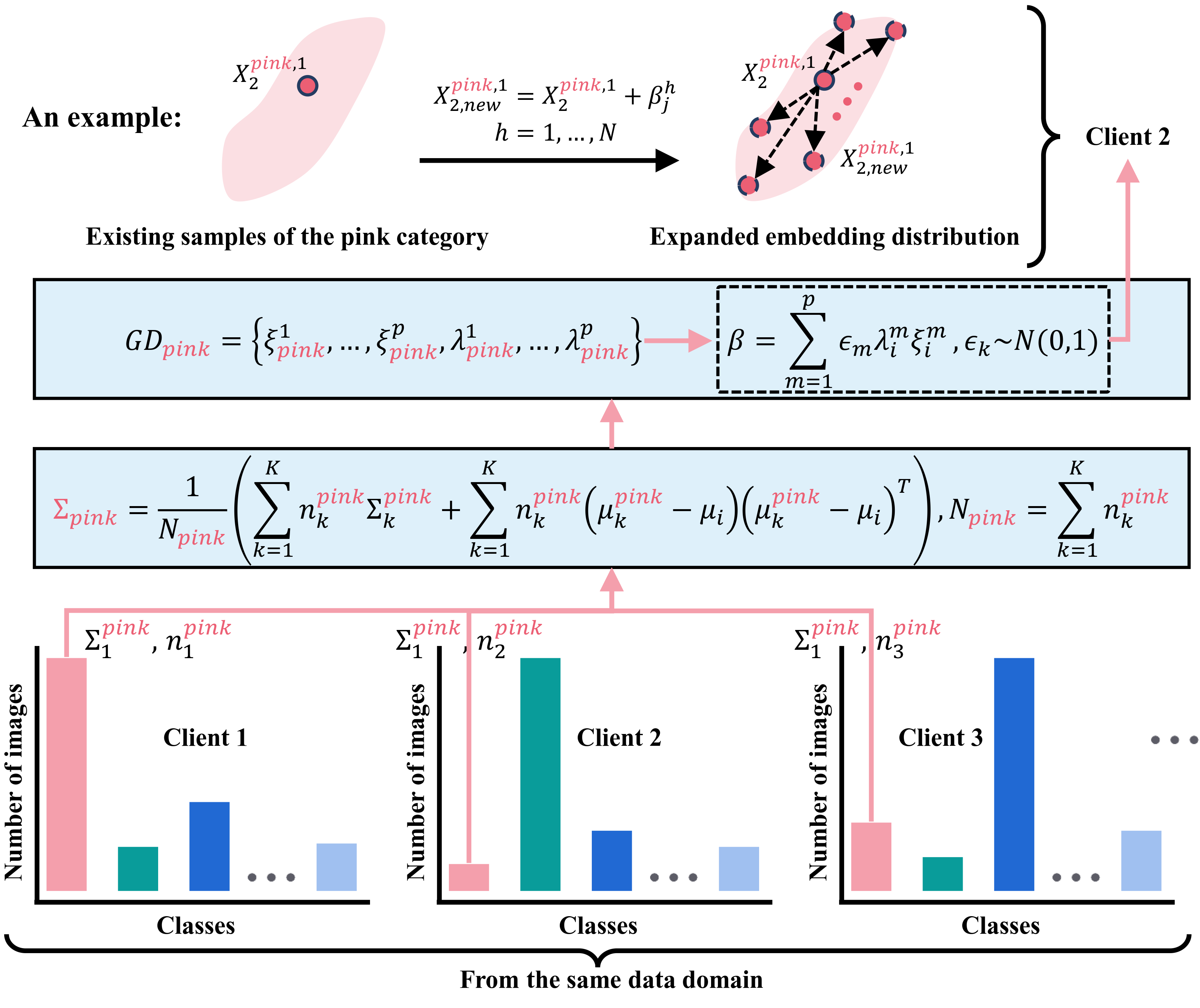}
\vskip -0.1in
   \caption{In a single-domain scenario, each class's global geometric shape is used to guide sample augmentation on each client. The example shows how new samples are generated for Client 2.}
   \label{fig7}
\end{figure}

We propose a Global Geometry-Guided Embedding Uncertainty Representation (\textbf{GGEUR)} for locally augmenting samples (i.e., embeddings) to simulate the global distribution. Specifically, for class \( i \), its global geometric shape can be represented as \( \text{GD}_i = \{\xi_i^1, \ldots, \xi_i^p, \lambda_i^1, \ldots, \lambda_i^p\} \). Suppose client \( k \) has only a limited number of class \( i \) embeddings, denoted as \( X_k^i = [X_k^{(i,1)}, \ldots, X_k^{(i,n_k^i)}] \in \mathbb{R}^{p \times n_k^i} \), where \( n_k^i \) represents the sample count. As shown in Fig.~\ref{fig7}, we first perform \( n_k^i \times N \) random linear combinations of the eigenvectors \( \xi_i^1, \ldots, \xi_i^p \) to produce \( n_k^i \times N \) different vectors $\beta = \sum_{m=1}^p \epsilon_m \lambda_i^m \xi_i^m \in \mathbb{R}^p$, where \( \epsilon_j \) follows a standard Gaussian distribution \( N(0, 1) \) and is scaled by the eigenvalues \( \lambda_i^j \) to control the magnitude. These vectors provide the direction and range for simulating the global distribution.
Next, for each existing sample \( X_k^{(i,1)}, \ldots, X_k^{(i,n_k^i)} \), we apply \( N \) of these vectors to generate \( n_k^i \times N \) new samples as follows:
\begin{equation}
\begin{split}
X_{(k,h)}^{(i,j)} = X_k^{(i,j)} + \beta_j^h,  j = 1, \ldots, n_k^i,  h = 1, \ldots, N.
\end{split}
\end{equation}
The total number of new samples can be set based on task requirements; in this study, for each local class, we ensure the total number of new and existing samples reaches $2000$. Fig.~\ref{fig7} and Algorithm \ref{alg1} illustrate this process in detail. After embedding space augmentation, each client trains an MLP as the local model.

\begin{algorithm}[t]
\caption{Global Geometry-Guided Embedding Uncertainty Representation (GGEUR)}
\label{alg1}
\textbf{Input:} $X_k^i = [X_k^{(i,1)}, \dots, X_k^{(i, n_k^i)}] \in \mathbb{R}^{p \times n_k^i}$: Sample set of class $i$ at client $k$, 
$GD_i = \{\xi_i^1, \dots, \xi_i^p, \lambda_i^1, \dots, \lambda_i^p\}$: Global geometric shape (eigenvectors and eigenvalues) of class $i$, 
$N$: Number of new samples to generate per original sample. \\
\textbf{Output:} New samples of class $i$.
\begin{algorithmic}[1] 
    \STATE $X_{\text{gen}} \gets \emptyset$ \# Initialize generated samples
    \FOR {$h = 1$ to $N$}
        \STATE $\beta^h \gets \sum_{m=1}^{p} \epsilon_m \lambda_i^m \xi_i^m, \epsilon_m \sim \mathcal{N}(0,1)$ 
        \STATE \# Generate new vector
        \STATE $X_{\text{gen}} \gets X_{\text{gen}} \cup \{X + \beta^h\}$ 
      \STATE \# Generate and add new sample, Equation (5)
    \ENDFOR
    \STATE \textbf{return} $X_{\text{gen}}$

\STATE $X_{\text{new}}^i \gets \emptyset$ \# Initialize augmented sample set
\FOR {$j = 1$ to $n_k^i$}
    \STATE $X_{\text{new}}^i \gets X_{\text{new}}^i \cup \textsc{GGEUR}(X_k^{(i,j)}, GD_i, N)$
\ENDFOR
\STATE \textbf{return} $X_{\text{new}}^i$
\end{algorithmic}
\end{algorithm}

\subsubsection{Multi-Domain Federated Learning}
\label{sec4.2.3}

\begin{figure}[t]
  \centering
   \includegraphics[width=1\linewidth]{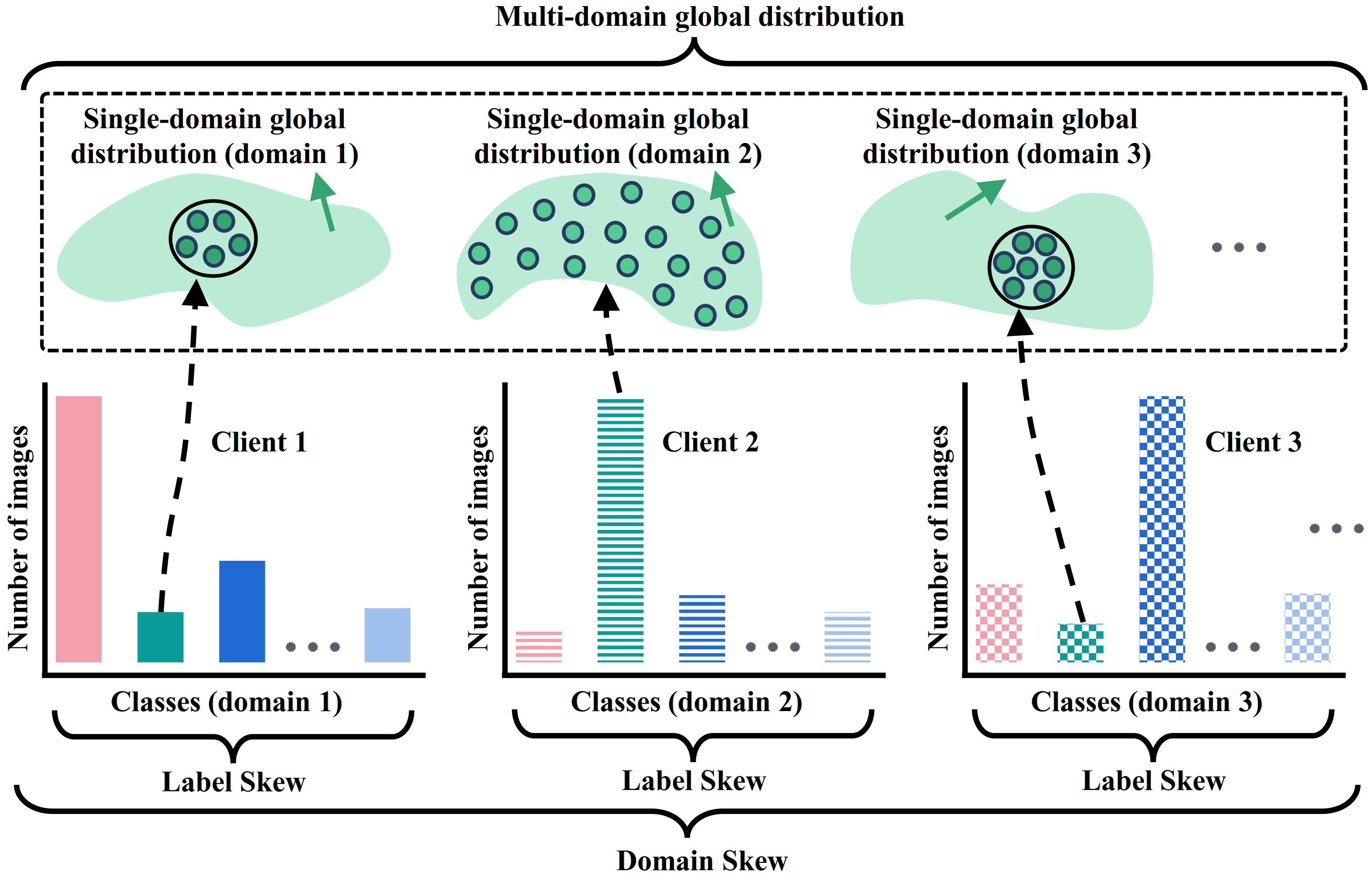}
\vskip -0.1in
   \caption{Federated scenario with both label skew and domain skew. Different textures represent data from distinct domains.}
   \label{fig8}
\vskip -0.1in
\end{figure}

When data originates from multiple domains, we encounter a complex scenario where both label skew and domain skew exist, as illustrated in Fig.~\ref{fig8}. In this case, different clients possess data from different domains. For clarity, we introduce two new concepts: the single-domain global distribution and the multi-domain global distribution.

\begin{definition}[\textbf{Single-Domain Global Distribution}]
Within a particular data domain, when a category’s sample count is sufficient and diverse enough to fully represent the category, this distribution of samples is termed the \textbf{single-domain global distribution} for that category. For example, in Fig.~\ref{fig8}, client $2$’s green samples are numerous and adequately cover the global distribution of the green category within domain $2$.
\end{definition}

\begin{definition}[\textbf{Multi-Domain Global Distribution}]
For a given category, the combined single-domain global distributions from all data domains constitute the \textbf{multi-domain global distribution} of that category. As shown in Fig.~\ref{fig8}, the green distributions across all domains together form the multi-domain global distribution for the green category.
\end{definition}

\begin{figure}[t!]
  \centering
   \includegraphics[width=1\linewidth]{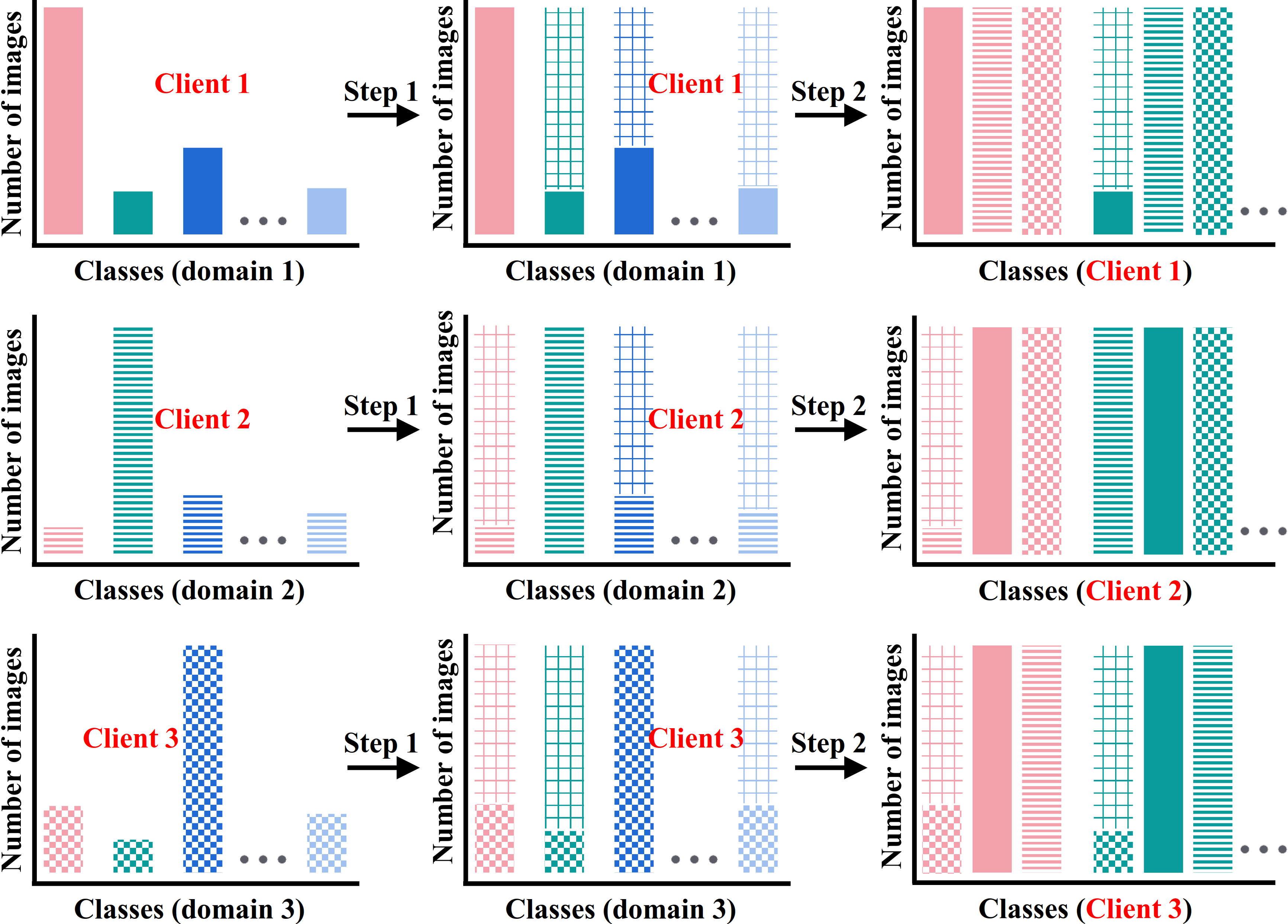}
\vskip -0.1in
   \caption{Example with 3 clients: Step 1 generates samples for each client from its own domain, while Step 2 simulates samples from other domains for each client.}
   \label{fig9}
\vskip -0.1in
\end{figure}

\begin{figure*}[!b]
  \centering
   \includegraphics[width=1\linewidth]{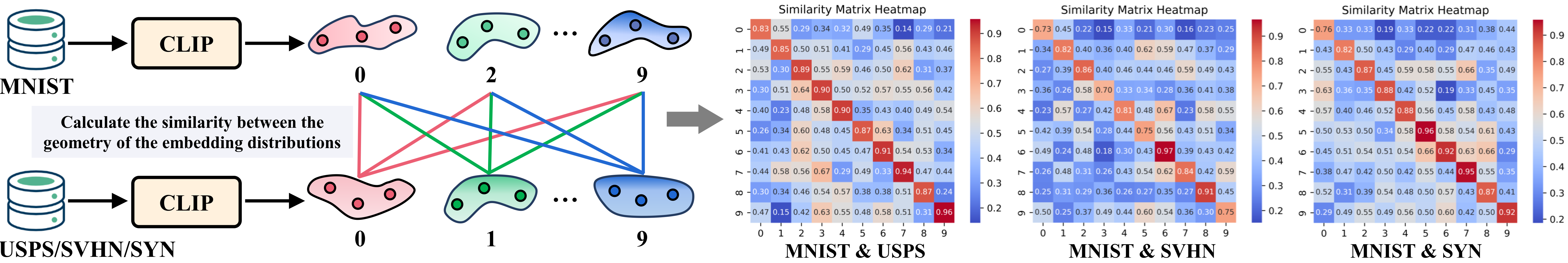}
\vskip -0.05in
   \caption{Geometric shape similarity between classes across domains on the Digits dataset.}
   \label{fig10}
\vskip -0.15in
\end{figure*}

Clearly, addressing both label skew and domain skew requires two steps: \textbf{(1)} Local sample augmentation to simulate the single-domain global distribution within each client’s data domain. For example, in Fig~\ref{fig8}, client 1 lacks sufficient green samples to represent the single-domain global distribution and thus requires sample augmentation. \textbf{(2)} Generation of new samples on each client to simulate the single-domain global distributions of other data domains.

As depicted in Fig.~\ref{fig9}, Step 1 addresses the label skew on each client, while Step 2 addresses domain skew. In summary, our objective is to simulate the ideal multi-domain global distribution for each category locally on each client. Unlike single-domain federated learning, completing Step 1 in this context poses the challenge of obtaining the geometric shape of the single-domain global distribution. For instance, in Figure \ref{fig9}, client 1 has only a small number of green samples, making it difficult to access the single-domain global distribution for the green category. Fortunately, when embeddings are extracted using CLIP, the geometric shapes of multiple single-domain global distributions corresponding to the same category across different domains are similar (Fig.~\ref{fig10}). This implies that we can identify a cross-domain shared geometric shape for each category, representing its global geometric shape.

Specifically, we investigate this using the Digits dataset, which includes four digit-recognition datasets (i.e., MNIST \cite{MNIST_IEEE98}, USPS \cite{USPS_PAMI94}, SVHN \cite{svhn_NeurIPS11}, and SYN \cite{syn}), each representing a different domain. First, we use CLIP (ViT-B/16) \cite{CLIP} to extract image embeddings for all categories across the four datasets. Then, we compute the similarity between the geometric shapes of the embedding distributions for each category in MNIST and the corresponding categories in USPS, SVHN, and SYN, as shown in Fig.\ref{fig10}. The significantly higher values in the diagonal of each heatmap indicate that, across domains, the geometric shapes of embeddings for the same category exhibit a consistent pattern, rather than being randomly distributed. This allows us to simplify the multi-domain label skew problem in Step 1 to a label skew problem in the single-domain scenario.

In detail, we first extract embeddings using CLIP and then locally calculate the covariance matrix and mean of local samples, which are uploaded to the server. Given CLIP's cross-domain consistency in representing the same class, we continue to use Equation (\textcolor{red}{10}) on the server to produce a shared global covariance matrix $\Sigma_1, \Sigma_2, \dots, \Sigma_C$ for each class. We further obtain the shared global geometric shapes, which are then sent to clients. Each client then applies GGEUR to augment samples, thereby simulating the single-domain global distribution locally. 
\textbf{Following this}, clients also need to generate new samples to simulate the single-domain global distribution from other data domains. Although the geometric shapes of embedding distributions across different domains are similar, the positions of these distributions differ. We propose transferring the class prototypes (local sample means) from other domains to the local client, and applying GGEUR to these prototypes to generate new samples, thus simulating the multi-domain global distribution. For each client, in Step 1, we ensure that the total number of new and existing samples for each class is $500$. Step 2 generates $500$ new samples based on each prototype separately.

\begin{figure*}[t]
\centering
\centerline{\includegraphics[width=2\columnwidth]{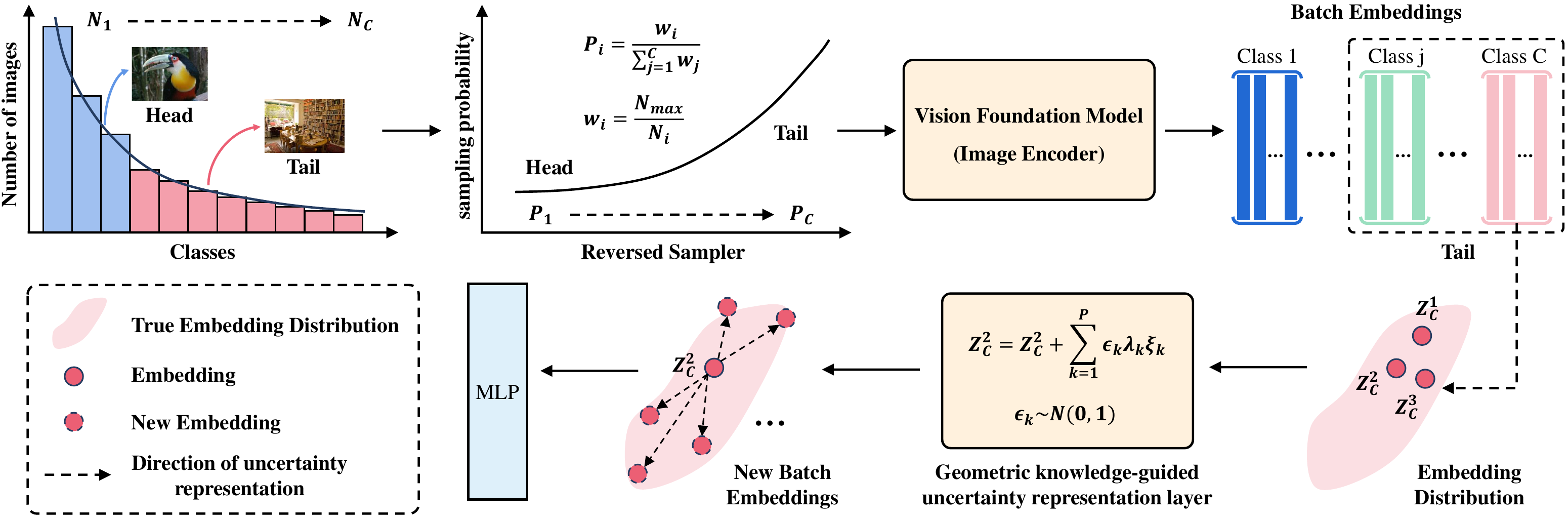}}
\caption{Complete Training Procedure Using GGEUR-Layer. At each iteration, inverse frequency sampling is conducted first, followed by passing the embeddings extracted by the foundation model to GGEUR-Layer, and finally, classification is performed by MLP.}
\label{fig11}
\end{figure*}

\subsection{Tail Class Distribution Calibration in Long-Tailed Image Recognition}
\label{sec4.3}

In this section, we first introduce how to adapt the \textbf{GGEUR} to calibrate the embedding distributions of tail classes (Section \ref{sec4.3.1}). Building upon this, we then propose a more concise geometric knowledge-based uncertainty representation layer for end-to-end training of long-tailed classifiers (Section \ref{sec4.3.2}). \textbf{It is worth noting that} our approach does not involve fine-tuning vision foundation models; it only requires calibration of embedding distributions to train a better long-tail classifier. Therefore, our method can also directly utilize pre-trained CLIP models to generate image embeddings of long-tailed distributions.

\subsubsection{Calibrating the Embedding Distribution of Tail Class}
\label{sec4.3.1}

Assuming ImageNet-1k is used as an external knowledge base, denoted by $\text{IN}_1,\dots,\text{IN}_{1000}$ representing $1000$ categories. Given a tail class $C_i$ in the long-tail dataset, the $p$-dimensional image embeddings of this class are extracted using a vision foundation model (CLIP/DINOv2) as $Z_i = [Z_i^1, \dots, Z_i^m] \in \mathbb{R}^{p \times m}$, where $m$ represents the number of samples. The prototype of tail class $C_i$ is computed as the mean of each dimension of the image embeddings: 
\begin{equation}
\begin{split}
\mu_i =( {\textstyle \sum_{j=1}^{m}Z_i^j})/m,
\end{split}
\end{equation}
and similarly for each category in ImageNet-1k.

Using cosine distance to measure inter-class similarity, let's assume that the class most similar to tail class $C_i$ in ImageNet-1k is \( \text{IN}_j \). We extract the image embeddings corresponding to \( \text{IN}_j \) using a vision foundation model as \( Z_{\text{IN}_j} = [Z_{\text{IN}_j}^1, \ldots, Z_{\text{IN}_j}^n] \in \mathbb{R}^{p \times n} \). The covariance matrix of the embedding distribution for class \( \text{IN}_j \) is estimated as:
\begin{equation}
\begin{split}
\Sigma_{\text{IN}_j} = \frac{1}{n} Z_{\text{IN}_j} \left(Z_{\text{IN}_j}\right)^T \in \mathbb{R}^{p \times p}.
\end{split}
\end{equation}
Performing eigenvalue decomposition on the matrix \( \Sigma_{\text{IN}_j} \) yields \( p \) eigenvalues \( \lambda_1 \geq \cdots \geq \lambda_P \) and their corresponding \( p \)-dimensional eigenvectors \( \xi_1, \ldots, \xi_p \). The eigenvectors and eigenvalues provide the direction and magnitude for augmenting and recovering the tail class distribution, respectively, which is guaranteed by cross-domain geometric consistency (Section \ref{sec2.2}).

Specifically, we aim to restore the ideal distribution of tail classes as much as possible by generating new samples for them in the embedding space. Firstly, we conduct $N$ random linear combinations of the eigenvectors \( \xi_1, \dots, \xi_P \) to obtain $N$ distinct vectors 
$
\beta = \sum_{k=1}^p \epsilon_k \lambda_k \xi_k \in \mathbb{R}^p,
$
where \( \epsilon_k \) follows the standard Gaussian distribution \( N(0,1) \). Next, using the existing samples \( Z_i^1 \) of tail class \( C_i \) as the center, we obtain $N$ new samples by $Z_i^1 + \beta$.
The same operation is applied to the remaining $m-1$ samples of tail class \( C_i \), resulting in a total of $N\times m$ new samples to restore the ideal distribution of tail class \( C_i \) as much as possible. In this work, we ensure that the number of samples for the augmented tail class is consistent with the number of samples for the most frequent class. For example, in CIFAR-10-LT, the number of samples for the augmented tail class is set to $5000$.

\subsubsection{A GGEUR-Based Layer for End-to-End Training}
\label{sec4.3.2}

Generating new samples is a direct way to help the tail class recover its ideal distribution, but it may not be very practical in engineering applications. This is because training the classifier end-to-end becomes impossible until calibration is performed on the tail class. Therefore, in the following, we propose a GGEUR-based Network layer (\textbf{GGEUR-Layer}), which not only achieves tail class calibration but also ensures that the model can be trained end-to-end.

Before training the long-tailed classifier using GGEUR-Layer, we pre-extracted the geometric shapes (including eigenvectors and eigenvalues) of each category in the external knowledge base using a vision foundation model and computed the embedding prototypes for each category. Similarly, we used the vision foundation model to extract feature embeddings for the tail classes in the long-tail dataset and calculated the embedding prototypes. We then matched each tail class to the most similar category in the knowledge base based on the cosine distance between the prototypes. This entire process is performed before using GGEUR-Layer, allowing us to select the already matched category's eigenvectors/eigenvalues for each tail class during the training of the long-tailed classifier.

Hereafter, we no longer generate additional samples for tail classes but instead learn the classifier directly from the long-tailed data. Therefore, to maintain a balanced optimization, we employ a mechanism of inverse sampling at each iteration \cite{bbn}. That is, if a class has more samples, its probability of being sampled is lower. Assuming there are $C$ classes, each with a sample count of $N_i$, the sampling probability for class $i$ can be calculated as
\begin{equation}
\setlength\abovedisplayskip{3pt} 
\setlength\belowdisplayskip{3pt}
\begin{split}
P_i = \frac{w_i}{\sum_{j=1}^{C} w_j}, \text{where}  \ w_i = \frac{N_{\text{max}}}{N_i}, \\ N_{\text{max}} = \max\{N_1, \ldots, N_C\}.
\end{split}
\end{equation}
Fig.~\ref{fig11} clearly illustrates the training process. A balanced mini-batch of training data is obtained through the reverse sampling process. However, in this batch, samples belonging to tail classes may be repeatedly sampled, lacking diversity. Therefore, we characterize each tail class sample in a batch with uncertainty representation to enhance information and calibrate tail classes. Specifically, given a tail class sample \( Z_C^i \), after passing through the GGEUR-Layer, \( Z_C^i \) is represented as a new embedding:
\begin{equation}
\setlength\abovedisplayskip{3pt} 
\setlength\belowdisplayskip{3pt}
\begin{split}
Z_C^i=Z_C^i+\sum_{k=1}^{P} \epsilon_k \lambda_k \xi_k, \text{where}  \ \epsilon_k \sim N(0,1).
\end{split}
\end{equation}
Finally, we employ a simple one-layer MLP to classify the long-tailed data, resulting in a very small number of learnable parameters. Since our method only calibrates the embedding distribution, it can be easily combined with other foundation models for long-tail recognition.

\section{Empirical Study}
\label{sec5}

This section aims to comprehensively evaluate the effectiveness, generality, and robustness of our proposed geometric knowledge-guided distribution calibration method. We design experiments covering two major scenarios: federated learning and long-tailed recognition, to demonstrate that the proposed method can effectively bridge the gap between local observations and the ideal global distribution.

\subsection{Datasets and Evaluation Metrics}
\label{sec5.1}

We conduct experiments using datasets for both federated learning and long-tailed recognition. The federated learning datasets are further categorized into three types: label skew, domain skew, and a combination of both.

\subsubsection{Federated Data Heterogeneity Scenarios}
\label{sec5.1.1}

\textbf{Label Skew Datasets.} We evaluate our method on three single-domain image classification tasks.
\begin{itemize}[]
    \item \textbf{Cifar-10} \cite{CIFAR} contains $10$ classes, with $50,000$ images for training and $10,000$ images for validation.
    \item \textbf{Cifar-100} \cite{CIFAR} covers $100$ classes, with $50,000$ training images and $10,000$ validation images.
\item \textbf{Tiny-ImageNet} \cite{tiny} is the subset of ImageNet with $100K$ images of size $64 \times 64$ with $200$ classes scale.
\end{itemize}

\noindent \textbf{Domain Skew Datasets.} We evaluated our method on the multi-domain image classification dataset Digits and conducted analyses on Office-Caltech and PACS.
\begin{itemize}[]
    \item \textbf{Digits} \cite{MNIST_IEEE98,USPS_PAMI94,svhn_NeurIPS11,syn} includes four domains: MNIST, USPS, SVHN and SYN, each with $10$ categories.
    \item \textbf{Office-Caltech} \cite{office_caltech} includes four domains: Caltech, Amazon, Webcam, and DSLR, each with $10$ categories.
\item \textbf{PACS} \cite{pacs} includes four domains: Photo (P) with $1, 670$ images, Art Painting (AP) with $2,048$ images, Cartoon (Ct) with $2,344$ images and Sketch (Sk) with $3,929$ images. Each domain holds seven categories.
\end{itemize}

\begin{figure}[t]
\centering
	\begin{minipage}{0.495\linewidth}
		\centering
		\includegraphics[width=1\linewidth]{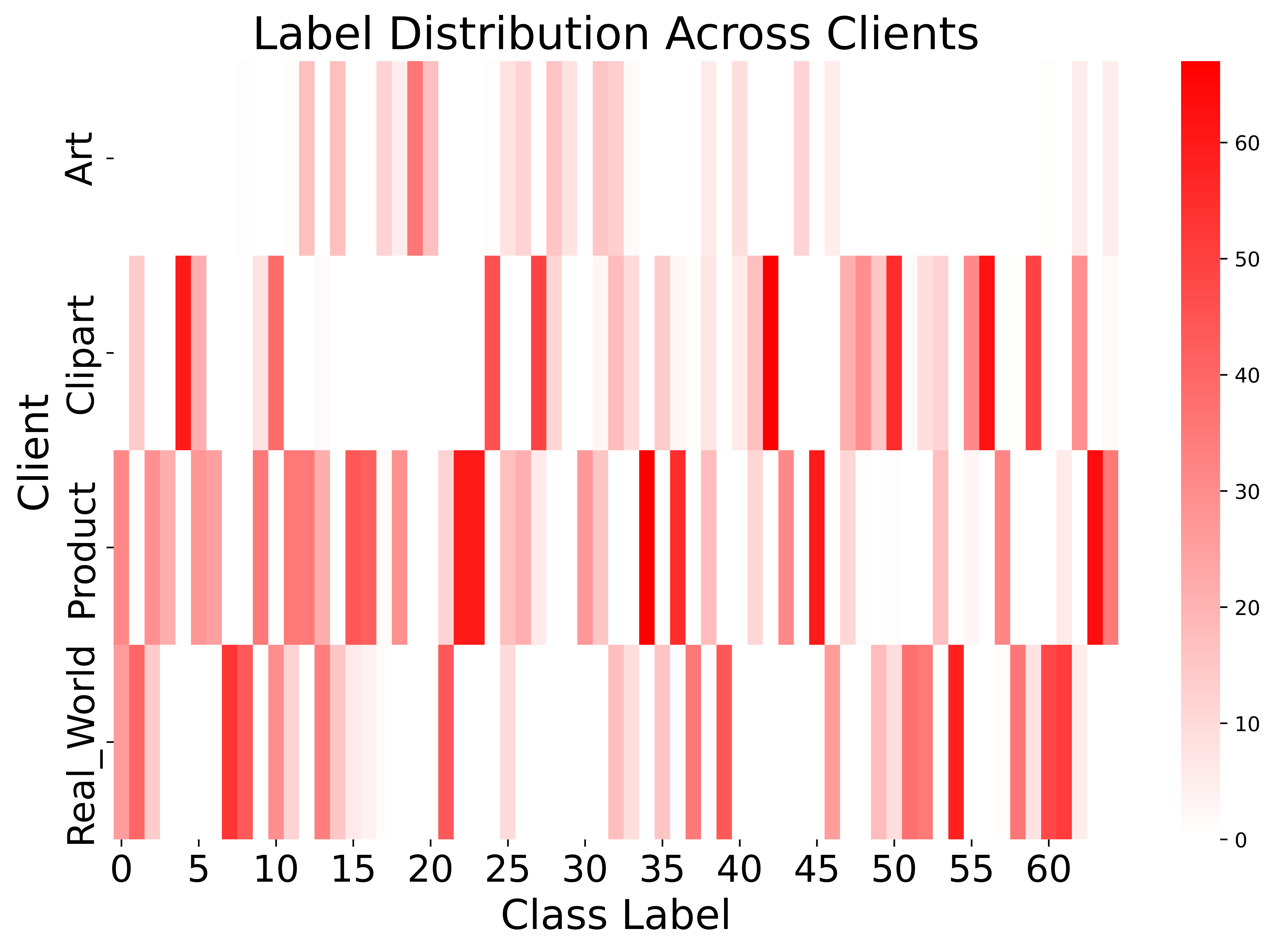}
	\end{minipage}
	\begin{minipage}{0.495\linewidth}
		\centering
		\includegraphics[width=1\linewidth]{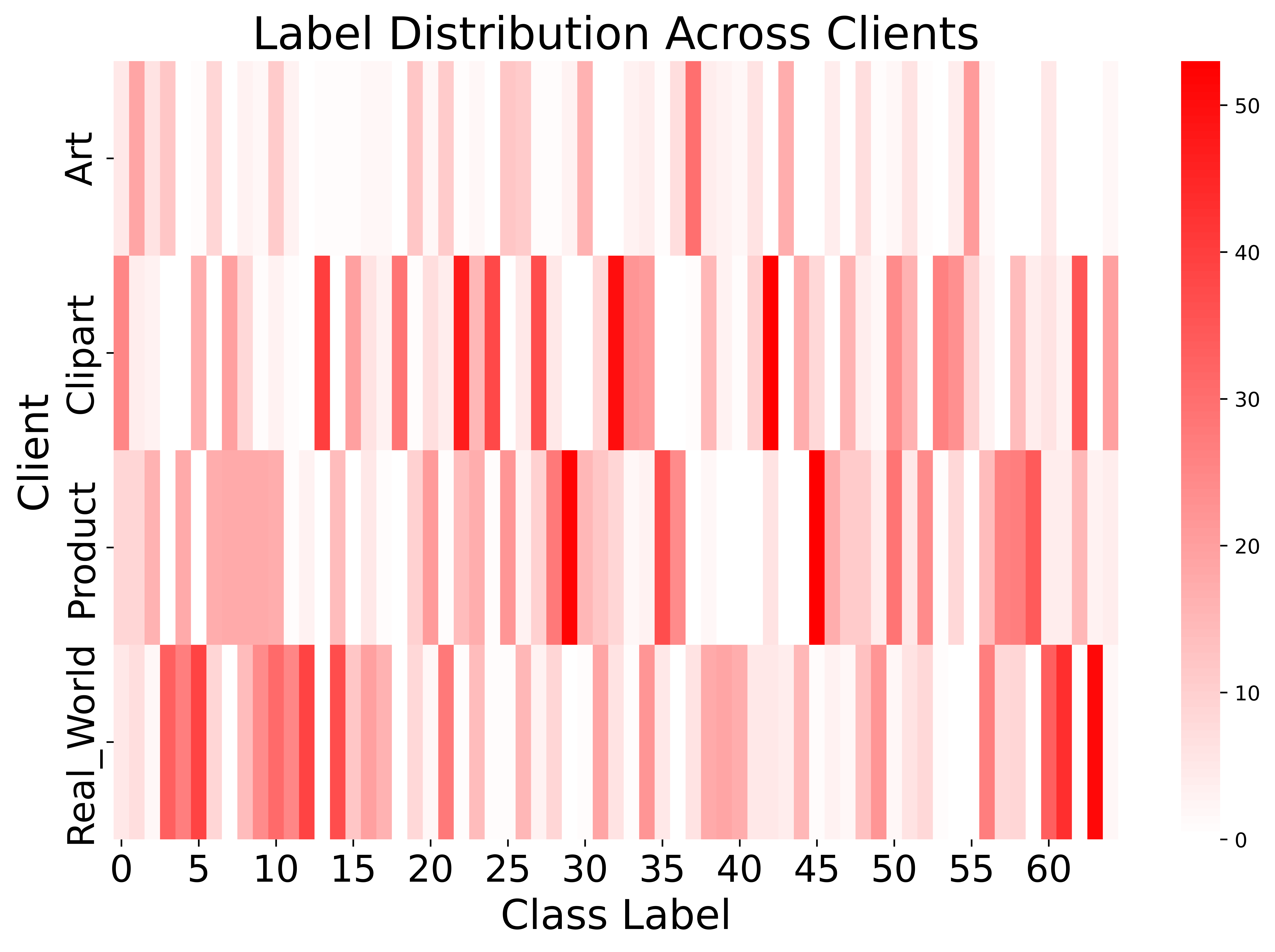}
	\end{minipage}
\caption{Number of samples per class across four clients when $\beta$ equals $0.1$ and $0.5$, with \textbf{each client holding data from a different domain.}}
\label{fig12}
\vskip -0.1in
\end{figure}

\noindent \textbf{Dataset with Coexisting Label Skew and Domain Skew.}
\textbf{Office-Home} \cite{office_home} includes $4$ domains: Art (A), Clipart (C), Product (P), and Real World (R), each containing $65$ classes. To increase the challenge, we designed a new partitioning method for the multi-domain dataset Office-Home to create a scenario where label skew and domain skew coexist. We name the newly constructed dataset \textbf{Office-Home-LDS (Label and Domain Skew)}. Figure \ref{fig9} shows the data distribution of Office-Home-LDS with different $\beta$ values. Dataset and Constructor published at: \url{https://huggingface.co/datasets/WeiDai-David/Office-Home-LDS}.


\subsubsection{Long-Tailed Recognition Datasets}
\label{sec5.1.2}

We evaluate our proposed method on four widely-adopted long-tailed benchmark datasets: CIFAR-10-LT, CIFAR-100-LT \cite{reb2}, ImageNet-LT \cite{LTdata}, and Places-LT \cite{LTdata}. The imbalance factor (IF) is defined as $\max_k\{n_k\}/\min_k\{n_k\}$, where $n_k$ is the number of samples in the $k$-th class.

\begin{itemize}[]
\item \textbf{CIFAR-10-LT / CIFAR-100-LT}: These are long-tailed versions of the CIFAR datasets \cite{CIFAR}, created by applying an exponential sampling scheme to the original balanced datasets. We conduct experiments with imbalance factors of $10$, $50$, $100$, and $200$.
    
\item \textbf{ImageNet-LT}: This is a long-tailed subset of the ILSVRC 2012 dataset \cite{LTdata}. It contains $115.8$K images from $1,000$ categories, with an imbalance factor of $256$. The number of images per class ranges from a maximum of $1,280$ to a minimum of $5$.
    
\item \textbf{Places-LT}: This is a long-tailed version of the Places-365 dataset \cite{LTdata}, containing $62.5$K images from $365$ classes. The number of images per class varies from $4,980$ to $5$, resulting in an imbalance factor of $100$.
\end{itemize}

\subsubsection{Evaluation Metrics}
\label{sec5.1.3}

\noindent\textbf{Federated Learning:} Following \cite{li2019fair}, we use the Top-1 Accuracy (\%) as the primary metric to evaluate the overall performance of the global model. To assess the fairness and stability of the model across different domains, we also report the Standard Deviation (STD) of the accuracy across all domains. A smaller standard deviation indicates better performance fairness and robustness to data heterogeneity. The final performance is reported as the average of the results from the last five communication rounds.

\noindent\textbf{Long-Tailed Recognition:} 
The Top-1 Accuracy (\%) on the test set is used as the primary performance metric. For ImageNet-LT and iNaturalist 2018, we also report the performance on the Head (classes with more than $100$ samples), Middle (classes with $20$ to $100$ samples), and Tail (classes with fewer than $20$ samples) subsets to provide a more comprehensive analysis.

\begin{table}[t]
\small
\setlength{\abovecaptionskip}{0cm}
\centering
\setlength{\tabcolsep}{2.4pt} 
\renewcommand\arraystretch{1.1}
\caption{Comparison results on CIFAR-100 and Tiny-ImageNet datasets with different degrees of label skew ($\beta$ values). The best results are shown in \underline{\textbf{underlined bold}}. FedAvg (CLIP+MLP) indicates that the backbone network uses CLIP and MLP, and the federated learning method employs FedAvg.}
\scalebox{0.96}{
\begin{tabular}{r||ccc|ccc}
\hline\thickhline
\rowcolor{lightgray}
& \multicolumn{3}{c}{CIFAR-100} & \multicolumn{3}{c}{Tiny-ImageNet} \\
\cline{2-7}
\rowcolor{lightgray}
\multirow{-2}{*}{Methods} 
 & 0.5 & 0.3 & 0.1  & 0.5 & 0.3 & 0.1  \\
\hline\hline

Zero-Shot CLIP & \multicolumn{3}{c|}{64.87}    & \multicolumn{3}{c}{63.67}   \\
FedTPG   & 71.40 & 70.95 & 68.63   & 67.63 & 66.72 & 64.71  \\
FedCLIP   & 72.03 & 71.20 & 70.64   & 70.41 & 70.37 & 69.50  \\
\hline \hline
FedMix (CLIP+MLP) & 81.31 & 79.62 & 73.85   & 70.89 & 68.57 & 63.43  \\
FEDGEN (CLIP+MLP) & 81.24 & 78.97 & 73.15   & 72.37 & 70.35 & 64.16  \\
FedFA (CLIP+MLP) & 81.98 & 79.31 & 74.68   & 70.41 & 70.68 & 64.62  \\ \hline 
FedAvg (CLIP+MLP)  & 81.41 & 77.68 & 68.22  & 70.08 &67.65  & 60.10  \\
\textbf{+ GGEUR}  & \underline{\textbf{83.31}} & \underline{\textbf{81.65}} & \underline{\textbf{77.70}}   & \underline{\textbf{73.89}} & \underline{\textbf{72.19}} & \underline{\textbf{66.86}}  \\
\bottomrule \hline
\end{tabular}}
\label{tab1}
\vskip -0.15in
\end{table}

\begin{table}[t]
\small
\setlength{\abovecaptionskip}{0cm}
\centering
\setlength{\tabcolsep}{4.1pt} 
\renewcommand\arraystretch{1.05}
\caption{Evaluation results of GGEUR on CIFAR-10 and CIFAR-100 with more severe label skew (i.e., smaller $\beta$ values).}
\scalebox{1}{
\begin{tabular}{r||ccccc}
\hline\thickhline
\rowcolor{lightgray}
& \multicolumn{5}{c}{CIFAR-10}  \\
\cline{2-6}
\rowcolor{lightgray}
\multirow{-2}{*}{Methods} 
 & 0.01 & 0.03 & 0.05  & 0.07 & 0.09   \\
\hline\hline
FedAvg (CLIP+MLP) & 90.87 & 90.13 & 91.96   &92.05 & 91.82 \\
\textbf{+ GGEUR}  & \underline{\textbf{94.39}} & \underline{\textbf{94.25}} & \underline{\textbf{95.07}}   & \underline{\textbf{95.21}} & \underline{\textbf{95.38}}  \\
\bottomrule 

\rowcolor{lightgray}
& \multicolumn{5}{c}{CIFAR-100}  \\
\cline{2-6}
\rowcolor{lightgray}
\multirow{-2}{*}{Methods} 
 & 0.01 & 0.03 & 0.05  & 0.07 & 0.09   \\
\hline\hline
FedAvg (CLIP+MLP) & 58.71 & 60.77 & 62.32   & 61.69 & 66.51 \\
\textbf{+ GGEUR}  & \underline{\textbf{75.72}} & \underline{\textbf{75.40}} & \underline{\textbf{75.96}}   & \underline{\textbf{76.72}} & \underline{\textbf{78.00}}  \\
\bottomrule \hline
\end{tabular}}
\label{tab2}
\vskip -0.15in
\end{table}

\begin{table}[t]
\small
\setlength{\abovecaptionskip}{0cm}
\centering
\setlength{\tabcolsep}{4pt} 
\renewcommand\arraystretch{1}
\caption{Evaluation results of GGEUR on Tiny-ImageNet with more severe label skew (i.e., smaller $\beta$ values).}
\scalebox{1}{
\begin{tabular}{r||ccccc}
\hline\thickhline
\rowcolor{lightgray}
& \multicolumn{5}{c}{Tiny-ImageNet}  \\
\cline{2-6}
\rowcolor{lightgray}
\multirow{-2}{*}{Methods} 
 &0.01  &0.03  & 0.05 &0.07 & 0.09     \\
\hline\hline
FedAvg (CLIP+MLP) & 53.03 & 54.57  & 58.91  &58.77 &59.13     \\
\textbf{+ GGEUR} & \underline{\textbf{64.27}}  & \underline{\textbf{65.79}}   & \underline{\textbf{66.49}} & \underline{\textbf{66.34}}  & \underline{\textbf{66.85}}   \\
\bottomrule \hline
\end{tabular}}
\label{tab3}
\vskip -0.1in
\end{table}

\subsection{Implementation Details and Comparison Methods}
\label{sec5.2}

\noindent\textbf{Federated Learning:} 
On the label skew dataset, we applied GGEUR to FedAvg \cite{fedavg} using CLIP (ViT-B/16) \cite{CLIP} as the backbone network and compared it with state-of-the-art federated data augmentation methods in heterogeneous federated learning, including FedMix \cite{yoon2021fedmix}, FEDGEN \cite{zhu2021data}, and FedFA \cite{zhou2023fedfa}. Additionally, we compared it with other advanced methods that use CLIP (ViT-B/16) as the backbone, such as FedTPG \cite{fedtpg} and FedCLIP \cite{fedclip}. On the domain skew and Office-Home-LDS datasets, we focused on exploring the enhancement effect of GGEUR across various federated architectures. Therefore, we applied GGEUR to FedAvg, SCAFFOLD \cite{SCAFFOLD}, MOON \cite{moon}, FedDyn \cite{acar2021federated}, FedOPT \cite{fedopt}, FedProto \cite{fedproto}, and FedNTD \cite{lee2022preservation}, all of which use CLIP for image feature extraction and \textbf{a single-layer MLP} as the local model. All experimental settings for FL methods are consistent with the latest benchmark \cite{fl3}.

\noindent\textbf{Long-Tailed Recognition:} 
We calibrate the embedding distributions extracted from various models (e.g., CLIP, BALLAD \cite{BALLAD}, DINOv2) using our GGEUR-Layer and train a single-layer MLP classifier. We use the SGD optimizer with a learning rate of $0.001$. The batch size is set to $64$ for CIFAR-10/100-LT (trained for $30$ epochs) and $1024$ for ImageNet-LT and Places-LT (trained for $10$ epochs). The geometric knowledge for tail class calibration is transferred from head classes within the same dataset (for ImageNet-LT) or from the ImageNet dataset as an external knowledge base (for other datasets).
We particularly focus on comparing knowledge transfer-based methods in the long-tailed recognition domain, including OFA \cite{OFA}, GistNet \cite{Gistnet}, CMO \cite{CMO}, FDC \cite{FDC}, H2T \cite{H2T}, and FUR \cite{FUR}. Additionally, we also compare with other state-of-the-art methods, including MiSLAS \cite{MiSLAS}, ResLT \cite{reslt}, and RIDE+CR \cite{CR}, as well as foundation model fine-tuning methods CoOp \cite{CoOp}, CLIP-Adapter \cite{Clip-adapter}, Tip-Adapter-F \cite{Tip-adapter}, LPT \cite{lpt}, Decoder and LIFT \cite{PEL}.

\begin{table}[t]
\small
\setlength{\abovecaptionskip}{0cm}
\centering
\setlength{\tabcolsep}{0.3pt} 
\renewcommand\arraystretch{1}
\caption{Evaluation results on the Digits dataset. GGEUR by default includes both Step 1 and Step 2.}
\scalebox{0.96}{
\begin{tabular}{r||cccc|cc}
\hline\thickhline
\rowcolor{lightgray}
& \multicolumn{6}{c}{Digits}  \\
\cline{2-7}
\rowcolor{lightgray}
\multirow{-2}{*}{Methods} 
 &MNIST  &USPS  &SVHN &SYN & AVG $\uparrow$ &STD $\downarrow$    \\
\hline\hline
FedMix (CLIP+MLP)  &95.03  &90.25 & 57.50 &72.60  &78.85  &14.89  \\
FEDGEN (CLIP+MLP) &95.85  &92.52  &58.77 &73.62  &80.19  &14.99  \\
FedFA (CLIP+MLP) &96.68  &92.97  &57.87  &75.53  &80.76  &15.44  \\ \hline  \hline 

FedAvg \cite{fedavg}  &90.40  &60.30  & 34.68 &46.99  & 58.09 & 20.74     \\
FedAvg (CLIP+MLP)  &95.12  &89.74  & 56.36 &65.17  & 76.60 &16.25   \\
\rowcolor{cvpryellow!10}
+ \textbf{GGEUR (Step 1)}  &96.02  &93.02  & 58.55 &73.13  & 80.18 &15.28   \\
\rowcolor{cvpryellow!25}
+ \textbf{GGEUR (Step 1 \& 2)}  &97.05  &94.12  &63.54 &74.73  & \underline{\textbf{82.36}} &\underline{\textbf{13.84}}   \\ \hline

SCAFFOLD \cite{SCAFFOLD} & 97.79 &94.45 & 26.64 & 90.69 & 77.39 &29.41   \\
SCAFFOLD (CLIP+MLP)  & 94.62 & 90.08 & 54.33 & 68.71 & 76.93 &16.31   \\
+ \textbf{GGEUR}  & 95.91 & 92.08 & 63.25 & 71.54 & \underline{\textbf{80.70}}  &\underline{\textbf{13.69}}  \\ \hline

MOON \cite{moon} & 92.78 & 68.11 & 33.36 & 39.28 & 58.36  &23.82   \\
MOON (CLIP+MLP)  & 75.64 & 73.09 & 38.83 & 52.74 & 60.07  &\underline{\textbf{15.14}}   \\
\rowcolor{cvpryellow!10}
\textbf{+ GGEUR (Step 1)}  &84.64  &81.96  &43.04  &60.35  & 67.50 &16.97   \\
\rowcolor{cvpryellow!25}
\textbf{+ GGEUR (Step 1 \& 2)} & 95.16 & 91.13 & 55.23 & 71.00 & \underline{\textbf{78.13}}  &16.08   \\ \hline

FedDyn \cite{acar2021federated} & 88.91 & 60.34 & 34.57 & 50.72 & 58.65   &19.76   \\
FedDyn (CLIP+MLP) & 95.46 & 92.13 & 58.89 & 70.30 & 79.19  &15.19  \\
\textbf{+ GGEUR}   & 97.07 & 94.02 & 63.34 & 74.83 & \underline{\textbf{82.31}}  &\underline{\textbf{13.88}} \\ \hline

FedOPT \cite{fedopt} & 92.71 & 87.62 & 31.32 & 87.92 & 74.89  &25.38   \\
FedOPT (CLIP+MLP)  & 94.57 & 88.79 & 58.65 & 66.47 & 77.12 &14.96  \\
\textbf{+ GGEUR}   & 96.43 & 93.47 & 62.35 & 70.75 & \underline{\textbf{80.75}}   &\underline{\textbf{14.54}}  \\ \hline

FedProto \cite{fedproto} & 90.54 & 89.54 & 34.61 & 58.00 & 68.18  &23.38   \\
FedProto (CLIP+MLP)  & 94.86 & 92.63 & 54.29 & 65.52 & 76.83  &17.40   \\
\textbf{+ GGEUR}   & 97.19 & 94.12 & 63.70 & 73.83 & \underline{\textbf{82.21}} &\underline{\textbf{13.96}}  \\ \hline

FedNTD \cite{lee2022preservation} & 52.31 & 58.07 & 18.03 & 97.29 & 56.43  &28.12  \\
FedNTD (CLIP+MLP)  & 95.82 & 91.43 & 58.26 & 69.95 & 78.86   &15.41  \\
\textbf{+ GGEUR}  & 97.08 & 94.32 & 63.57 & 73.53 & \underline{\textbf{82.13}} &\underline{\textbf{14.06}}   \\ 

\bottomrule \hline
\end{tabular}}
\label{tab4}
\vskip -0.1in
\end{table}

\begin{table}[t]
\small
\setlength{\abovecaptionskip}{0cm}
\centering
\setlength{\tabcolsep}{1.15pt} 
\renewcommand\arraystretch{1}
\caption{Evaluation results on Office-Caltech using the CLIP+MLP model.}
\scalebox{0.96}{
\begin{tabular}{r||cccc|cc}
\hline\thickhline
\rowcolor{lightgray}
& \multicolumn{6}{c}{Office-Caltech}  \\
\cline{2-7}
\rowcolor{lightgray}
\multirow{-2}{*}{Methods} 
 &Am  &Ca  &D &W & AVG $\uparrow$ &STD $\downarrow$    \\
\hline\hline
FedAvg \cite{fedavg}  & 81.99 & 73.21 & 79.37 & 67.93 & 75.62 & 6.31     \\
FedAvg (CLIP+MLP)  & 98.26 & 96.74 & 100 & 100 & \underline{\textbf{98.75}} &\underline{\textbf{1.57}}   \\ \hline

SCAFFOLD \cite{SCAFFOLD} & 39.77 & 42.50 & 78.02 & 70.69 & 57.75 &19.44   \\
SCAFFOLD (CLIP+MLP)  & 96.18 & 92.88 & 95.83 & 93.26 & \underline{\textbf{94.54}} &\underline{\textbf{1.70}}   \\ \hline

MOON \cite{moon} & 84.42 & 75.98 & 84.67 & 68.97 & 78.51  &7.53   \\
MOON (CLIP+MLP)  & 98.61 & 97.33 & 100 &98.88 & \underline{\textbf{98.70}}  &\underline{\textbf{1.09}}   \\ \hline

FedDyn \cite{acar2021federated} & 84.02 & 72.59 & 77.34 & 68.97 & 75.72   &6.50   \\
FedDyn (CLIP+MLP) & 98.61 & 96.74 & 100 & 100 & \underline{\textbf{98.84}}  &\underline{\textbf{1.54}}  \\  \hline

FedOPT \cite{fedopt}  & 79.05 & 71.96 & 89.34 & 74.48 & 78.71  &7.67   \\
FedOPT (CLIP+MLP)  & 98.26 & 97.33 & 100 & 100 &\underline{\textbf{ 98.90}} &\underline{\textbf{1.33}}  \\  \hline

FedProto \cite{fedproto} & 87.79 & 75.98 & 90.00 & 79.31 & 83.27  &6.70   \\
FedProto (CLIP+MLP)  & 98.26 & 96.74 & 100 & 100 & \underline{\textbf{98.75}}  &\underline{\textbf{1.57}}   \\   \hline

FedNTD \cite{lee2022preservation} & 10.95 & 10.89 & 14.67 & 10.34 & 11.71  &1.99  \\
FedNTD (CLIP+MLP)  & 97.92 & 96.14 & 100 & 100 & \underline{\textbf{98.51}}  &\underline{\textbf{1.86}}  \\

\bottomrule \hline
\end{tabular}}
\label{tab5}
\vskip -0.1in
\end{table}

\begin{table}[t]
\small
\setlength{\abovecaptionskip}{0cm}
\centering
\setlength{\tabcolsep}{1.1pt} 
\renewcommand\arraystretch{1}
\caption{Evaluation results on PACS using the CLIP+MLP model.}
\scalebox{0.96}{
\begin{tabular}{r||cccc|cc}
\hline\thickhline
\rowcolor{lightgray}
& \multicolumn{6}{c}{PACS}  \\
\cline{2-7}
\rowcolor{lightgray}
\multirow{-2}{*}{Methods} 
 &P  &AP  &Ct &Sk & AVG $\uparrow$ &STD $\downarrow$    \\
\hline\hline
FedAvg \cite{fedavg}  & 76.09 & 64.19 & 83.50 & 89.40 & 78.30 & 9.41     \\
FedAvg (CLIP+MLP)  & 99.40 & 98.37 & 99.01 & 93.64 &\underline{\textbf{97.60}}  &\underline{\textbf{2.32}}   \\ \hline

SCAFFOLD \cite{SCAFFOLD} & 61.95 & 45.44 & 58.87 & 54.64 & 55.25 &6.22   \\
SCAFFOLD (CLIP+MLP)  & 92.42 & 81.63 & 80.68 & 87.28 & \underline{\textbf{85.50}} &\underline{\textbf{4.72}}   \\ \hline

MOON \cite{moon} & 74.44 & 64.19 & 83.92 & 89.17 & 77.93  &9.53   \\
MOON (CLIP+MLP)  & 99.60 & 99.02 & 99.43 & 93.89 &\underline{\textbf{97.99}}  &\underline{\textbf{2.37}}   \\ \hline

FedDyn \cite{acar2021federated} & 78.17 & 63.29 & 82.27 & 89.93 & 78.66  &9.70   \\
FedDyn (CLIP+MLP) & 99.40 & 98.37 & 99.01 & 93.55 & \underline{\textbf{97.58}}   &\underline{\textbf{2.36}}  \\  \hline

FedOPT \cite{fedopt} & 78.66 & 67.66 & 82.41 & 83.68 & 78.12 &6.31   \\
FedOPT (CLIP+MLP)  & 99.40 & 98.37 & 99.01 & 93.64 & \underline{\textbf{97.60}}  &\underline{\textbf{2.32}}  \\  \hline

FedProto \cite{fedproto}  & 85.63 & 73.69 & 83.57 & 91.14 & 83.51  &6.31   \\
FedProto (CLIP+MLP)  & 99.40 & 98.21 & 99.01 & 94.23 & \underline{\textbf{97.71}}  &\underline{\textbf{2.06}}   \\   \hline

FedNTD \cite{lee2022preservation}  & 16.77 & 18.23 & 28.47 & 93.18 & 39.16  &31.51  \\
FedNTD (CLIP+MLP)  & 99.40 & 98.54 & 99.29 & 92.28 & \underline{\textbf{97.38}}  &\underline{\textbf{2.96}}  \\

\bottomrule \hline
\end{tabular}}
\label{tab6}
\vskip -0.1in
\end{table}

\begin{table}[t]
\small
\setlength{\abovecaptionskip}{0cm}
\centering
\setlength{\tabcolsep}{1pt} 
\renewcommand\arraystretch{1.1}
\caption{Evaluation results on Office-Home-LDS ($\beta=0.1$).}
\scalebox{0.91}{
\begin{tabular}{r||cccc|cc}
\hline\thickhline
\rowcolor{lightgray}
& \multicolumn{6}{c}{Office-Home-LDS}  \\
\cline{2-7}
\rowcolor{lightgray}
\multirow{-2}{*}{Methods} 
 &A  &C  &P &R & AVG $\uparrow$ &STD $\downarrow$    \\
\hline\hline
FedAvg (CLIP+MLP) \cite{fedavg}  &65.29 & 58.17 & 80.56 & 76.53 & 70.14 & 8.89     \\
\textbf{+ GGEUR}  &78.33 & 79.01 & 90.17 & 88.46 & \underline{\textbf{83.99}}   &\underline{\textbf{5.36}}   \\ \hline


SCAFFOLD (CLIP+MLP) \cite{SCAFFOLD} & 68.72 & 66.79 & 83.63 & 80.12 & 74.82 &7.20   \\
\textbf{+ GGEUR}  & 78.60 & 78.32 & 89.86 & 89.07 & \underline{\textbf{83.96}} &\underline{\textbf{5.51}}   \\ \hline

MOON (CLIP+MLP) \cite{moon} &69.27 & 68.63 & 86.56 & 82.87 & 76.83 &7.99   \\
\textbf{+ GGEUR}  & 72.02 & 70.31 & 86.11 & 83.87 & \underline{\textbf{78.08}}  &\underline{\textbf{6.98}}   \\ \hline

FedDyn (CLIP+MLP) \cite{acar2021federated} &58.30 & 55.19 & 77.63 & 72.86 & 65.99 &9.47   \\
\textbf{+ GGEUR} &78.88 & 78.55 & 90.47 & 88.46 & \underline{\textbf{84.09}}  &\underline{\textbf{5.42}}  \\  \hline

FedOPT (CLIP+MLP) \cite{fedopt} &58.44 & 54.89 & 76.80 & 72.25 & 65.59  &9.16   \\
\textbf{+ GGEUR}  & 79.01 & 78.32 & 90.84 & 88.61 & \underline{\textbf{84.20}}  &\underline{\textbf{5.59}}  \\  \hline

FedProto (CLIP+MLP) \cite{fedproto}  & 65.84 & 56.49 & 80.41 & 74.85 & 69.40   &9.09   \\
\textbf{+ GGEUR}  &78.05 & 77.71 & 89.79 & 87.84 & \underline{\textbf{83.35}}  &\underline{\textbf{5.51}}   \\   \hline

FedNTD (CLIP+MLP) \cite{lee2022preservation}  &69.68 & 66.64 & 84.53 & 80.96 & 75.46 &7.48  \\
\textbf{+ GGEUR}  &78.19 & 74.66 & 90.24 & 86.77 & \underline{\textbf{82.46}}  &\underline{\textbf{6.29}}  \\

\bottomrule \hline
\end{tabular}}
\label{tab7}
\vskip -0.1in
\end{table}

\subsection{Evaluation on Federated Learning Datasets}
\label{sec5.3}

\subsubsection{Evaluation Results on Label Skew Dataset}
\label{sec5.3.1}

\noindent \textbf{Main Results.} Tables \ref{tab1}, \ref{tab2}, and \ref{tab3} show the performance improvement of GGEUR over FedAvg (CLIP+MLP) \cite{fedavg} under different $\beta$ values, along with comparison results with other methods. Under all label distribution skew settings, GGEUR significantly outperforms other methods, demonstrating superior classification accuracy and adaptability to imbalanced label distributions. Particularly at lower $\beta$ values (i.e., with more severe label skew), GGEUR notably enhances the performance of FedAvg (CLIP+MLP) on CIFAR-100, validating its effectiveness in handling extreme label skew scenarios. For example, when $\beta$ is $0.01$, $0.03$, and $0.05$, GGEUR achieves performance gains of $\textbf{17.01\%}$, $\textbf{14.63\%}$, and $\textbf{13.64\%}$, respectively. These results indicate that GGEUR not only exhibits good generalization ability in standard settings but also maintains robust performance in cases of significant label skew. 

\noindent \textbf{Comparison with Peer Methods.} We implemented FedMix, FEDGEN, and FedFA for comparison in Table \ref{tab1}, and in all cases, our method outperformed these three approaches. This demonstrates the superiority of geometric shapes as knowledge over the Gaussian assumption and traditional data augmentation methods.

\subsubsection{Evaluation Results on Domain Skew Dataset}
\label{sec5.3.2}

\noindent \textbf{Ablation Study.} As shown in Fig.~\ref{fig9}, simulating the multi-domain global distribution requires two steps. We selected the classic FedAvg (CLIP+MLP) and MOON (CLIP+MLP) for an ablation study on the Digits dataset. The experimental results, \textcolor{cvpryellow}{highlighted in yellow} in Table \ref{tab4}, show that each step incrementally improves the original methods, and their combination achieves the best results.

\noindent \textbf{Main Results.} Tables \ref{tab4}, \ref{tab5}, and \ref{tab6} present the experimental results on datasets with domain skews. It can be observed that simply using CLIP \cite{CLIP} for image representation, combined with a single-layer MLP for federated learning, already surpasses existing methods. This improvement is attributed to the advancements in the foundation model, and GGEUR can further significantly enhance the performance of the global model. For instance, on the Digits dataset, GGEUR improves the average performance of FedAvg \cite{fedavg}, MOON \cite{moon}, and FedProto \cite{fedproto} by $\textbf{5.76\%}$, $\textbf{18.06\%}$, and $\textbf{5.38\%}$, respectively. Additionally, compared to other methods, GGEUR significantly reduces the accuracy variance across different domains. This demonstrates that GGEUR effectively adapts to features from different domains when handling cross-domain data distribution disparities, providing more robust and fair performance. 
On the Office-Caltech and PACS datasets, since the use of CLIP+MLP as the backbone network already achieves very high performance across all methods, these datasets are less challenging. However, our evaluation results can serve as a reference for other research.

\noindent \textbf{Comparison with Peer Methods.} We implemented FedMix (CLIP+MLP), FEDGEN (CLIP+MLP), and FedFA (CLIP+MLP) for comparison in Table \ref{tab4}. When GGEUR is applied to FedAvg (CLIP+MLP), it outperforms these three methods by $\textbf{3.51\%}$, $\textbf{2.17\%}$, and $\textbf{1.60\%}$, respectively.

\begin{figure}[t]
\centering
	\begin{minipage}{0.495\linewidth}
		\centering
		\includegraphics[width=1\linewidth]{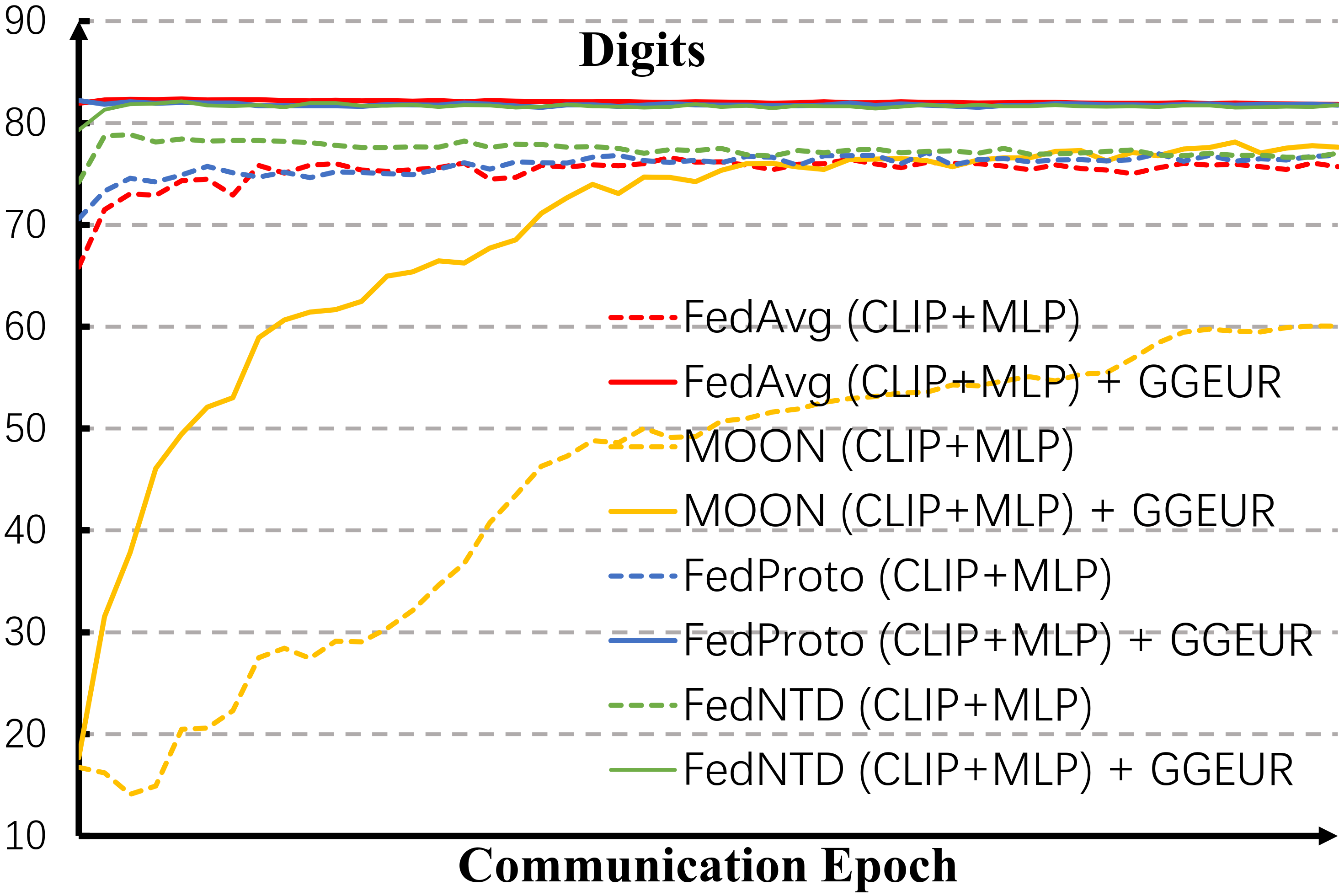}
	\end{minipage}
	\begin{minipage}{0.495\linewidth}
		\centering
		\includegraphics[width=1\linewidth]{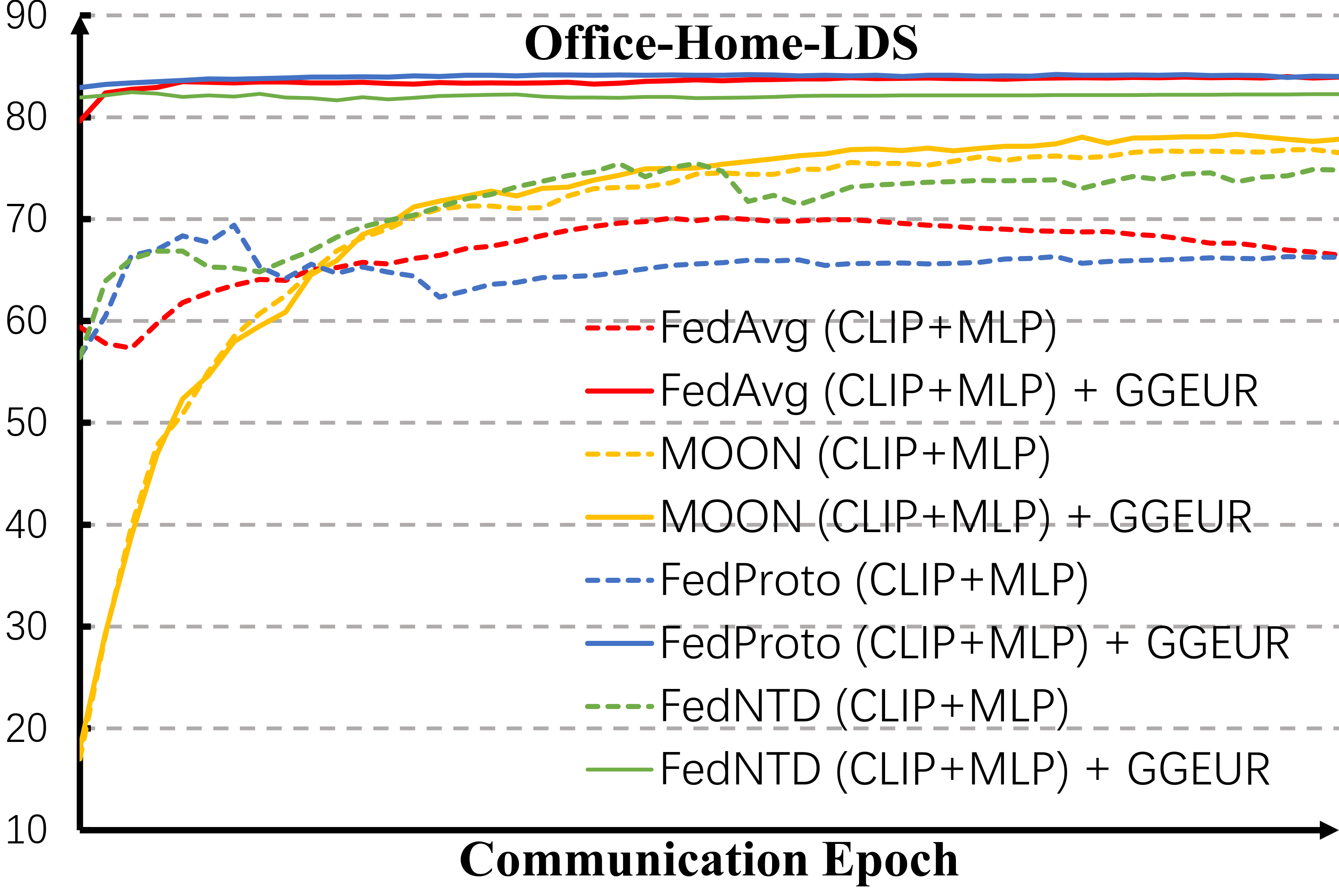}
	\end{minipage}
\vskip -0.05in
\caption{Comparison of convergence in average accuracy with and without integrating GGEUR in the selected FL methods.}
\label{fig13}
\vskip -0.1in
\end{figure}

\subsubsection{Evaluation Results on Office-Home-LDS}
\label{sec5.3.3}

To fully demonstrate the potential of GGEUR, we constructed a more challenging dataset, Office-Home-LDS, which incorporates both label skew and domain skew. Office-Home-LDS simulates a more realistic and complex scenario of distributional imbalance, encompassing cross-domain data distribution differences and label imbalance. In Table \ref{tab7}, we show the remarkable performance gains of GGEUR over existing methods on this dataset, along with significant reductions in accuracy variance across different domains, highlighting its effectiveness in handling highly heterogeneous data scenarios. Specifically, GGEUR improved the average performance of FedAvg \cite{fedavg}, SCAFFOLD \cite{SCAFFOLD}, FedDyn \cite{acar2021federated}, FedOPT \cite{fedopt}, FedProto \cite{fedproto}, and FedNTD \cite{lee2022preservation} by $\textbf{13.85\%}$, $\textbf{9.14\%}$, $\textbf{18.1\%}$, $\textbf{18.61\%}$, $\textbf{13.95\%}$, and $\textbf{7.0\%}$, respectively. These results demonstrate that GGEUR can achieve robust performance on complex multi-domain, multi-class datasets, making it an effective approach for addressing both label and domain skew.

\begin{table}[t]
\small
\setlength{\abovecaptionskip}{0cm}
\centering
\setlength{\tabcolsep}{0.5pt} 
\renewcommand\arraystretch{1}
\caption{Number of Clients $K$ Impact on Performance.}
\scalebox{1}{
\begin{tabular}{r||ccc}
\hline\thickhline
\rowcolor{lightgray}
& \multicolumn{3}{c}{CIFAR-10 ($\beta = 0.1$)}  \\
\cline{2-4}
\rowcolor{lightgray}
\multirow{-2}{*}{Methods} 
 &$K=100$  &$K=300$  &$K=500$    \\
\hline\hline

FedAvg (CLIP+MLP) & 87.89 & 84.69  & 82.05      \\
+ \textbf{GGEUR} & 93.55 ($+5.66$)  &92.17 ($+7.84$)   &90.43 ($+8.38$)   \\
\bottomrule \hline
\end{tabular}}
\label{tab8}
\end{table}

\begin{table}[t]
\vskip -0.1in
\small
\setlength{\abovecaptionskip}{0cm}
\centering
\setlength{\tabcolsep}{8pt} 
\renewcommand\arraystretch{1}
\caption{The average training time (s) per round.}
\scalebox{1}{
\begin{tabular}{r||ccc}
\hline\thickhline
\rowcolor{lightgray}
& \multicolumn{3}{c}{Digits}  \\
\cline{2-4}
\rowcolor{lightgray}
\multirow{-2}{*}{Methods} 
 &FedAvg  &SCAFFOLD  &MOON    \\
\hline\hline

CLIP+MLP & 28.2 & 54.5  & 32.3      \\
+ \textbf{GGEUR} & 31.7 (+3.5)  &59.1 (+4.6)   &35.6 (+3.3)   \\
\bottomrule \hline
\end{tabular}}
\label{tab9}
\end{table}

\begin{table*}[!b]
\caption{Comparison on CIFAR-10-LT and CIFAR-100-LT. The accuracy (\%) of Top-1 is reported. The best and second-best results are shown in \underline{\textbf{underlined bold}} and \textbf{bold}, respectively.}
\label{tab10}
\vskip -0.1in
\centering  
\begin{footnotesize}
\renewcommand\arraystretch{0.9}
\setlength{\tabcolsep}{15pt} 
\begin{tabular}{l|c|c|cccccc}
\hline \toprule
Dataset      &Backbone    & Pub.    & \multicolumn{3}{c|}{CIFAR-10-LT}               & \multicolumn{3}{c}{CIFAR-100-LT} \\ \hline
Imbalance Factor (IF) &\multicolumn{1}{c|}{-}  &\multicolumn{1}{c|}{-}  & 200  & 100   & \multicolumn{1}{c|}{50}   & 200    & 100    & 50      \\ \bottomrule
Cross Entropy   & ResNet-32  &\multicolumn{1}{c|}{-}  & 65.6 & 70.3 & 74.8  & 34.8   & 38.2   & 43.8    \\ \bottomrule
\multicolumn{9}{l}{\textbf{State-of-the-art long-tail knowledge transfer methods}}  \\ \bottomrule

OFA \cite{OFA}      & ResNet-32       & ECCV 2020    & 75.5 & 82.0 & \multicolumn{1}{c|}{84.4} & 41.4   & 48.5   & 52.1    \\
CMO \cite{CMO}    & ResNet-32         & CVPR 2022    &\multicolumn{1}{c}{-}            &\multicolumn{1}{c}{-}      & \multicolumn{1}{c|}{-}     &\multicolumn{1}{c}{-}        & 50.0   & 53.0    \\ 
FDC \cite{FDC} & ResNet-32  & TMM 2024    &79.7      &83.4     &\multicolumn{1}{c|}{86.5}     &45.8       & 50.6   & 54.1     \\ 
GCL+H2T \cite{H2T} & ResNet-32  & AAAI 2025   &79.4      &82.4          & \multicolumn{1}{c|}{85.4}         &45.2       & 48.9   & 53.8   \\ 
FUR \cite{FUR}   & ResNet-32    &IJCV 2025         & 79.8 &83.7  & \multicolumn{1}{c|}{86.2}  &46.2   &50.9  &54.1    \\   \bottomrule

\multicolumn{9}{l}{\textbf{Other state-of-the-art methods}}  \\ \bottomrule

MiSLAS \cite{MiSLAS}    & ResNet-32      & CVPR 2021    &\multicolumn{1}{c}{-}      & 82.1  & \multicolumn{1}{c|}{85.7} &\multicolumn{1}{c}{-}        & 47.0   & 52.3     \\ 


RIDE (4*) + CR  \cite{CR}   & ResNet-32            & CVPR 2023    & \multicolumn{1}{c}{-}  & \multicolumn{1}{c}{-}  & \multicolumn{1}{c|}{-}     &\multicolumn{1}{c}{-}   & 49.8   & 59.8      \\ 
RIDE + H2T \cite{H2T}   & ResNet-32            & AAAI 2025    & \multicolumn{1}{c}{-}  & \multicolumn{1}{c}{-}  & \multicolumn{1}{c|}{-}  &46.6   & 51.4   & 55.5         \\  \bottomrule 

\multicolumn{9}{l}{\textbf{Fine-tuning foundation model}}  \\ \bottomrule

BALLAD  \cite{BALLAD}   &ViT-B/16           &  \multicolumn{1}{c|}{-}    & \multicolumn{1}{c}{-}  & \multicolumn{1}{c}{-}  & \multicolumn{1}{c|}{-}     &\multicolumn{1}{c}{-}   & 77.8   & \multicolumn{1}{c}{-}   \\ 

CLIP (Zero-Shot) \cite{CLIP}   & ViT-B/16     & ICML 2022    &73.8   &73.8   & \multicolumn{1}{c|}{73.8}     &52.2   &52.2    & 52.2         \\


CoOp  \cite{CoOp}    & ViT-B/16           & IJCV 2022    & 74.4  & 76.1 & 78.6     &54.3   & 54.6   & 57.8         \\

CLIP-Adapter  \cite{Clip-adapter}    & ViT-B/16           & IJCV 2024    & 72.4  &75.6  & \multicolumn{1}{c|}{79.7}    &58.9  &61.7   & 62.5     \\  

LIFT \cite{PEL}               & ViT-B/16           &ICML 2024       & \multicolumn{1}{c}{-}  & \multicolumn{1}{c}{-}  & \multicolumn{1}{c|}{-}     &\multicolumn{1}{c}{-}   & \textbf{80.3}   &\textbf{82.0}      \\  \bottomrule   

\multicolumn{9}{l}{\textbf{Calibrating embedding distributions (Ours)}}  \\ \bottomrule
CLIP + MLP \cite{CLIP}   & ViT-B/16     & ICML 2023    &82.4      &84.7    &\multicolumn{1}{c|}{88.5}  &47.5        & 49.6   & 51.4     \\ 

+ GGEUR-Layer   & ViT-B/16     & \multicolumn{1}{c|}{-}    &\textbf{94.4}      &\textbf{94.6}    &\multicolumn{1}{c|}{\textbf{94.8}}  &\textbf{74.5}        & 78.9   & 81.6     \\ \hline

DINOv2 + MLP \cite{DINOv2}     & ViT-B/16           & TMLR 2024    & 90.3  & 92.1  & \multicolumn{1}{c|}{93.4}     &70.7   & 76.2  & 79.2       \\

+ GGEUR-Layer     & ViT-B/16           & \multicolumn{1}{c|}{-}      &\underline{\textbf{96.9}}   &\underline{\textbf{97.3}} & \multicolumn{1}{c|}{\underline{\textbf{97.5}}}     &\underline{\textbf{79.9}}   & \underline{\textbf{83.5}}   & \underline{\textbf{86.7}}       \\  \bottomrule  \hline

\end{tabular}
\end{footnotesize}
\end{table*}

\subsubsection{GGEUR Accelerates Convergence}
\label{sec5.3.4}

In Figure \ref{fig13}, we plot the average accuracy per epoch for various FL methods with and without using GGEUR. It can be observed that our method enhances the convergence speed of the FL methods, resulting in smoother curves.

\subsubsection{Large-Scale Client and Computational Cost}
\label{sec5.3.5}

In practical federated learning settings, the number of participating clients can significantly affect model performance due to increased data heterogeneity. To further evaluate the performance of our proposed method under larger-scale federated learning scenarios, we conducted additional experiments with an increased number of clients. Specifically, we conducted experiments on the label-skewed dataset CIFAR-10 with $100$, $300$, and $500$ clients. As shown in Table \ref{tab8}, the results demonstrate that GGEUR remains robust and continues to enhance the performance of FedAvg.

We conducted experiments on the domain-skewed dataset Digits, comparing the training time required to complete the full model for FedAvg, SCAFFOLD, and MOON before and after applying GGEUR. The results (Table \ref{tab9}) show that GGEUR introduces almost no additional training time overhead. Specifically, after applying GGEUR, the training time for the three methods increased by only $3.5s$, $4.6s$, and $3.3s$, respectively.

\begin{table*}[!t]
\caption{Comparisons (Top-1 accuracy (\%)) with state-of-the-art methods on ImageNet-LT and Places-LT. The best and the second-best results are shown in \underline{\textbf{underline bold}} and \textbf{bold}, respectively.}
\label{tab11}
\vskip -0.1in
\centering  
\begin{footnotesize}
\renewcommand\arraystretch{0.8}
\setlength{\tabcolsep}{10.5pt} 
\begin{tabular}{l|c|cccc|cccc}
\hline \toprule
\multirow{3}{*}{Methods} & \multirow{3}{*}{Pub.} & \multicolumn{4}{c|}{ImageNet-LT} & \multicolumn{4}{c}{Places-LT} \\  \cmidrule(lr){3-10}
                         &                       & Head  & Middle  & Tail & Overall & Head  & Middle  & Tail & Overall \\ \bottomrule

\multicolumn{10}{l}{\textbf{State-of-the-art long-tail knowledge transfer methods}}  \\ \bottomrule

                         &                       & \multicolumn{4}{c|}{ResNext-50}  & \multicolumn{4}{c}{ResNet-152}    \\  \hline

OFA \cite{OFA}                   & ECCV 2020          & 47.3  & 31.6    & 14.7 & 35.2    & 42.8      &37.5    &22.7   &36.4     \\
GistNet \cite{Gistnet}              & ICCV 2021              & 52.8  & 39.8    & 21.7 & 42.2    &42.5       &40.8        & 32.1    &39.6     \\

RIDE + CMO* \cite{CMO}                & CVPR 2022                & 66.4  & 53.9    & 35.6 & 56.2    & \multicolumn{1}{c}{-}  & \multicolumn{1}{c}{-}      & \multicolumn{1}{c}{-}  & \multicolumn{1}{c}{-}      \\ 
RIDE + H2T  \cite{H2T}                   & AAAI 2025               &67.6      &54.9         &37.1      &56.9    &43.0      &42.6         &36.3      &41.4   \\  



FUR \cite{FUR}             &IJCV 2025  & 65.4  & 52.2    & 37.8 &55.5    & \multicolumn{1}{c}{-}    & \multicolumn{1}{c}{-}     &\multicolumn{1}{c}{-}   & \multicolumn{1}{c}{-}      \\  \hline

\multicolumn{10}{l}{\textbf{Other state-of-the-art methods}}  \\ \bottomrule

MiSLAS \cite{MiSLAS}                 & CVPR 2021                  & 62.5  & 49.8   & 34.7 & 52.3    &39.6  & 43.3    &36.1 & 40.4    \\

DSB + RIDE \cite{DSB}                   & ICLR 2023               &68.6      &54.5        &38.5      &58.2    & \multicolumn{1}{c}{-}  & \multicolumn{1}{c}{-}      & \multicolumn{1}{c}{-}  & \multicolumn{1}{c}{-}   \\
ResLT \cite{reslt}                   & TPAMI 2023               &59.4     &51.0        &41.3      &52.7    &40.3       &44.4         &34.7      &41.0 \\
RIDE (4*) + CR  \cite{CR}                   & CVPR 2023               &68.5      &54.2         &38.8      &57.8    &\multicolumn{1}{c}{-}        &\multicolumn{1}{c}{-}         &\multicolumn{1}{c}{-}       &\multicolumn{1}{c}{-}   \\           \bottomrule

\multicolumn{10}{l}{\textbf{Fine-tuning foundation model}}  \\ \bottomrule
                         &                       & \multicolumn{4}{c|}{ViT-B/16}  & \multicolumn{4}{c}{ViT-B/16}    \\  \hline

CLIP (Zero-Shot) \cite{CLIP}                   & ICML 2022              & 67.7  & 66.5    & 66.4 & 67.0    &34.7   &37.9     &44.7  &39.2    \\
CoOp \cite{CoOp}                   & IJCV 2022            & 74.6  & 68.4    & 65.6 & 70.4    &41.8   &38.5    &44.3 &40.9     \\ 
Tip-Adapter-F   \cite{Tip-adapter}                   &ECCV 2022              & 74.2  & 73.2    & 61.1 & 71.8    &38.3   &45.1    &33.4  &40.2     \\
LPT \cite{lpt}                   & ICLR 2023            & \multicolumn{1}{c}{-}  & \multicolumn{1}{c}{-}    & \multicolumn{1}{c}{-} &   \multicolumn{1}{c|}{-}    & 49.3   &52.3     &46.9  &50.1     \\ 
Decoder \cite{Decoder}            & IJCV 2024            &\multicolumn{1}{c}{-}   &\multicolumn{1}{c}{-}    &\multicolumn{1}{c}{-}    & 73.2    &\multicolumn{1}{c}{-}   &\multicolumn{1}{c}{-}    &\multicolumn{1}{c}{-}   &46.8     \\  
LIFT \cite{PEL}             & ICML 2024            & 80.2    &\textbf{76.1}    & \textbf{71.5}    & \textbf{77.0}    &\textbf{51.3}   &\textbf{52.2}   &\textbf{50.5}   &\textbf{51.5}     \\  \bottomrule 

\multicolumn{10}{l}{\textbf{Calibrating embedding distributions (Ours)}}  \\ \bottomrule

CLIP + MLP \cite{CLIP}                & ICML 2023           &\underline{\textbf{84.5}}   & 56.8    & 35.7 & 64.6    & \underline{\textbf{51.4}}  & 31.6    &21.3  &36.5     \\
+ GUR               &\multicolumn{1}{c|}{-}                     &\textbf{80.6}  & 71.9    & 60.4 & 73.5    &42.9   &40.6     &44.5  &42.1     \\   \hline

DINOv2 + MLP \cite{DINOv2}            &TMLR 2024           &80.2   & 68.4    & 52.6 & 70.3    &40.6  & 41.0    &33.4  &39.3     \\
+ GGEUR-Layer               &\multicolumn{1}{c|}{-}                     &80.3  & 75.2    & 69.1 & 76.5    &45.2   &43.8     &42.5  &44.3     \\   \hline

BALLAD   \cite{BALLAD}                   &   \multicolumn{1}{c|}{-}           & 79.1  & 74.5    &69.8 &75.7    &49.3   &50.2    &48.4  &49.5    \\
+ GGEUR-Layer                    &    \multicolumn{1}{c|}{-}             & 80.5  & \underline{\textbf{77.8}}    &\underline{\textbf{74.6}} & \underline{\textbf{78.5}}    &51.0  & \underline{\textbf{52.6}}  &\underline{\textbf{51.2}}   & \underline{\textbf{51.9}}    \\ \bottomrule \hline
\end{tabular}
\end{footnotesize}
\vskip -0.1in
\end{table*}

\begin{figure*}[t]
\centering
\centerline{\includegraphics[width=2\columnwidth]{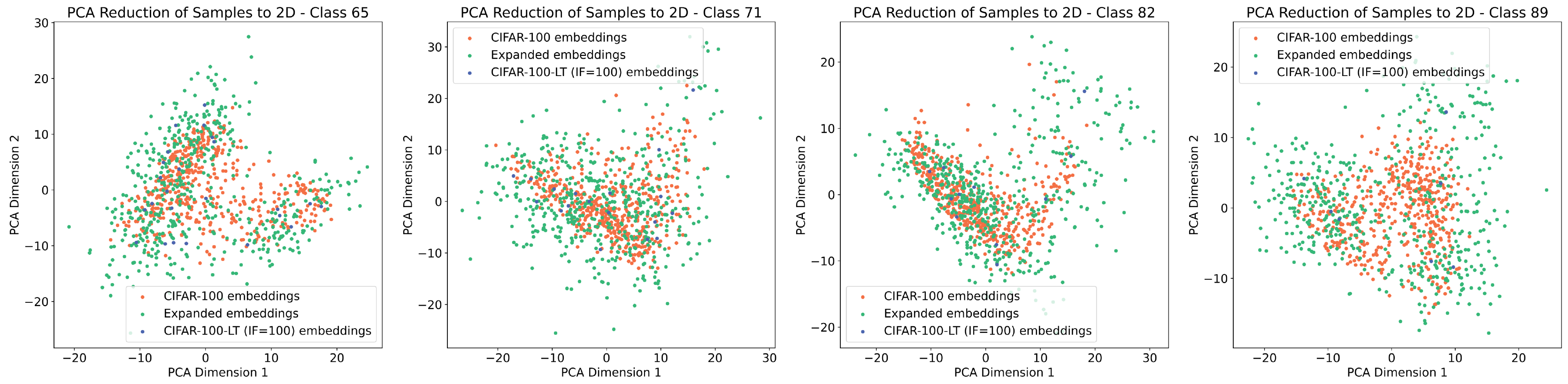}}
\vskip -0.1in
\caption{The Calibration Effectiveness of Tail Class Embedding Distributions.}
\label{fig14}
\vskip -0.05in
\end{figure*}

\subsection{Evaluation on Long-Tailed Datasets}
\label{sec5.4}

\subsubsection{Results on CIFAR-10-LT and CIFAR-100-LT}
\label{sec5.4.1}

The experimental results are summarized in Table \ref{tab10}. Our method has leaped forward on CIFAR-10-LT and CIFAR-100-LT. Particularly in CIFAR-10-LT, the performance of DINOv2+MLP+GUR surpasses existing state-of-the-art methods at different IF settings: by $\textbf{17.1\%}$ over FUR at IF $200$, $\textbf{13.6\%}$ over FUR at IF $100$, and $\textbf{11\%}$ over FDC at IF $50$. Similarly, in CIFAR-100-LT, at IF of $200$, $100$, and $50$, DINOv2+MLP+GUR outperforms the leading long-tailed recognition method CLIP-Adapter by $\textbf{21\%}$, $\textbf{21.8\%}$, and $\textbf{24.2\%}$, respectively. This significant performance enhancement stems not only from the exceptional performance of the foundation models themselves but also from the outstanding enhancing effect of GGEUR-Layer in long-tailed scenarios, which makes our approach markedly superior to base model fine-tuning methods. For instance, GGEUR-Layer enables CLIP+MLP to achieve performances of $\textbf{27\%}$, $\textbf{29.3\%}$, and $\textbf{30.2\%}$ on CIFAR-100-LT at different IF settings.

\subsubsection{Results on ImageNet-LT and Places-LT}
\label{sec5.4.2}

Table \ref{tab11} demonstrates the significant enhancement effect of GGEUR-Layer on CLIP and BALLAD. On the Tail subsets of ImageNet-LT and Places-LT, GGEUR-Layer enables CLIP to achieve performance gains of $\textbf{24.7\%}$ and $\textbf{23.2\%}$ respectively. Even though BALLAD is specifically designed for long-tailed scenarios, GGEUR-Layer still improves the overall performance of BALLAD by $\textbf{2.8\%}$ and $\textbf{2.4\%}$ on these two datasets. We observe that GGEUR-Layer sometimes reduces the performance of the Head subset, but its ability to significantly improve the performance of the Middle and Tail subsets leads to a smaller bias while enhancing the overall performance of the model. 
We would like to explain this from the perspective of IGAM \cite{IGAM}. Since the MLP in CLIP+MLP is trained directly on long-tailed data, the resulting decision space is pathological. Specifically, IGAM visualized the weight norms of each class in classifiers trained on long-tail data and observed that the weight norms were highly imbalanced, leading to a pathological decision boundary where the decision space for tail classes is severely compressed. In summary, the good performance of CLIP+MLP on head classes comes at the cost of severely impairing the performance of middle and tail classes. For example, CLIP+MLP achieves $51.4\%$ accuracy on head classes in Places-LT but only $21.3\%$ accuracy on tail classes.
The success of GGEUR-Layer is attributed to both the strong capability of the foundation model and several phenomena discovered exclusively on the base model. The collision and fusion of the foundation model with prior knowledge make our method significantly superior to existing methods.

\begin{figure}[t] 
\begin{center}
\includegraphics[width=1\columnwidth]{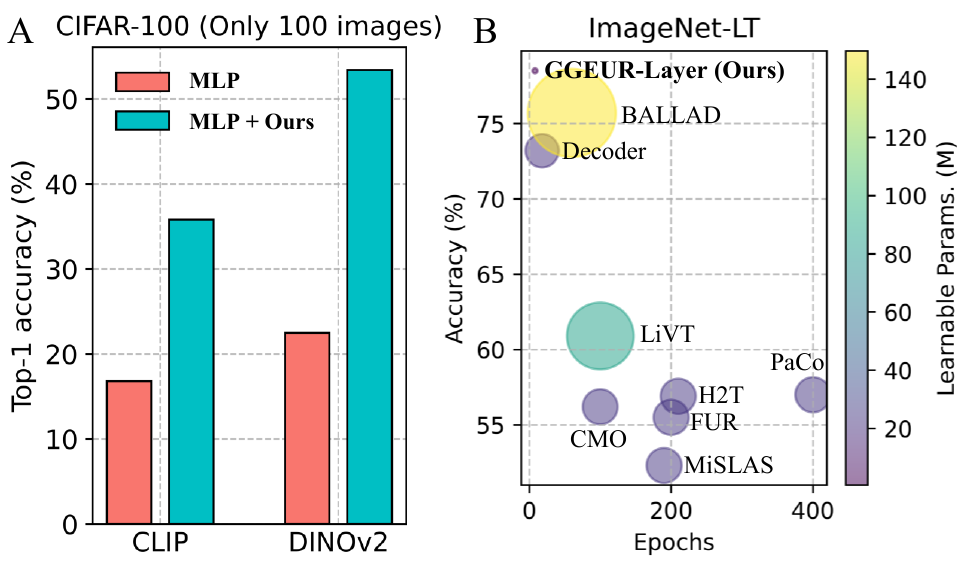}
\vskip -0.15in
\caption{A. Performance of GGEUR-Layer in extreme scenarios. B. Number of Learnable Parameters and Training Speed.}
\vskip -0.2in
\label{fig15}
\end{center}
\end{figure}

\subsection{More Advantages and Analysis}
\label{sec5.4.3}

\textbf{Visualization Examples of Tail Class Calibration.} We utilize our method to generate new samples for the tail classes on CIFAR-100 with an IF of $100$ and visualize the results. As shown in Fig.~\ref{fig14}, the green samples generated from a small number of blue samples cover the real distribution (i.e., orange samples) well. It is particularly noteworthy that the geometric shape of the new distribution is very close to that of the real distribution, which strongly validates the rationality of our motivation.

\textbf{An extreme example.} Randomly select one image from each class of CIFAR-100, totaling 100 images. Extract embeddings for the 100 images using CLIP and DINOv2, then compare the performance of the MLP trained before and after using GGEUR-Layer on the test set. Fig.~\ref{fig15}A illustrates that GGEUR-Layer still plays a significant role.

\textbf{Fewer Learnable Parameters (M) and Faster Training Speed.} We compared our method with traditional approaches and fine-tuning methods based on the foundation model. As depicted in Fig.~\ref{fig15}B, our approach demonstrates superior performance while requiring fewer learnable parameters and converging faster.

\section{Conclusion}
\label{sec:conclusion}

This work proposes a novel paradigm for geometric knowledge guided distribution calibration to address the fundamental challenge of distribution missing under data scarcity. We reveal that the geometric structure of class distributions in vision foundation models exhibits strong cross-domain consistency, serving as a transferable deep prior. Leveraging this insight, we unify distribution calibration in federated learning and long-tailed recognition by treating geometry as guiding knowledge. Our approach shifts the focus from conventional optimization compensation and data augmentation to knowledge transfer and distribution reconstruction, enabling models to effectively "imagine" and calibrate unobserved distributions. The framework not only offers a new technical pathway for data-constrained scenarios but also provides insights for broader applications such as few-shot learning and domain generalization.

\section*{Acknowledgments}
This work was in part supported by National Natural Science Foundation of China (Grant Nos. 92370201 and 62222607).

\bibliographystyle{IEEEtran}
\bibliography{nips2024}

\vspace{-1cm}
\begin{IEEEbiography}[{\includegraphics[width=0.9in,height=1.25in,clip,keepaspectratio]{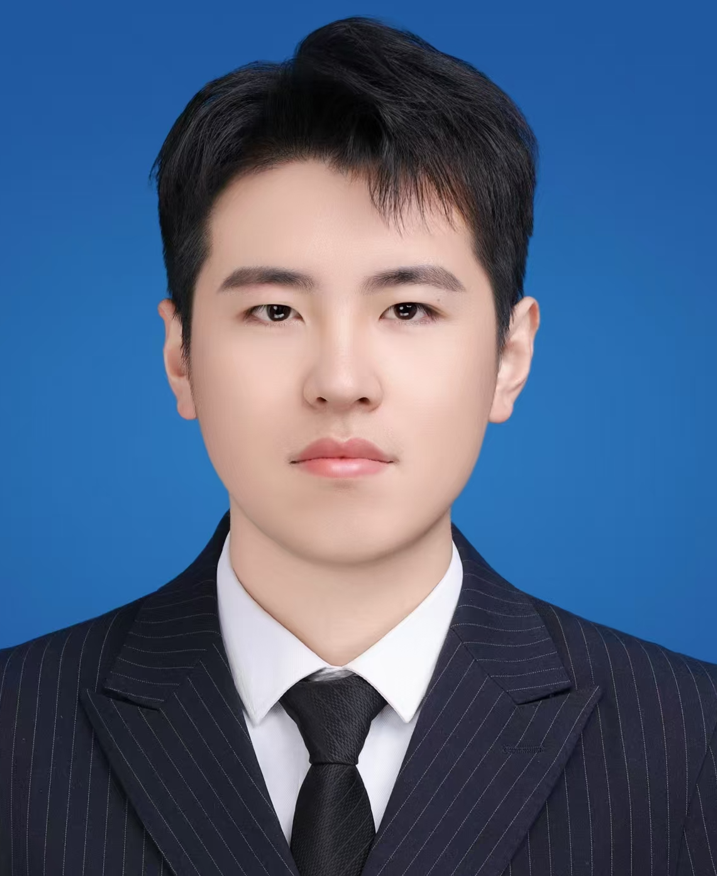}}]{Yanbiao Ma}
received his B.Eng. in Intelligent Science and Technology from Xidian University in 2020 and his Ph.D. in Computer Science and Technology from Xidian University in 2025. He is currently an Assistant Professor at the Gaoling School of Artificial Intelligence, Renmin University of China. His research focuses on imbalanced learning and fairness in deep neural networks.
\end{IEEEbiography}

\vspace{-1.5cm}
\begin{IEEEbiography}[{\includegraphics[width=0.9in,height=1.25in,clip,keepaspectratio]{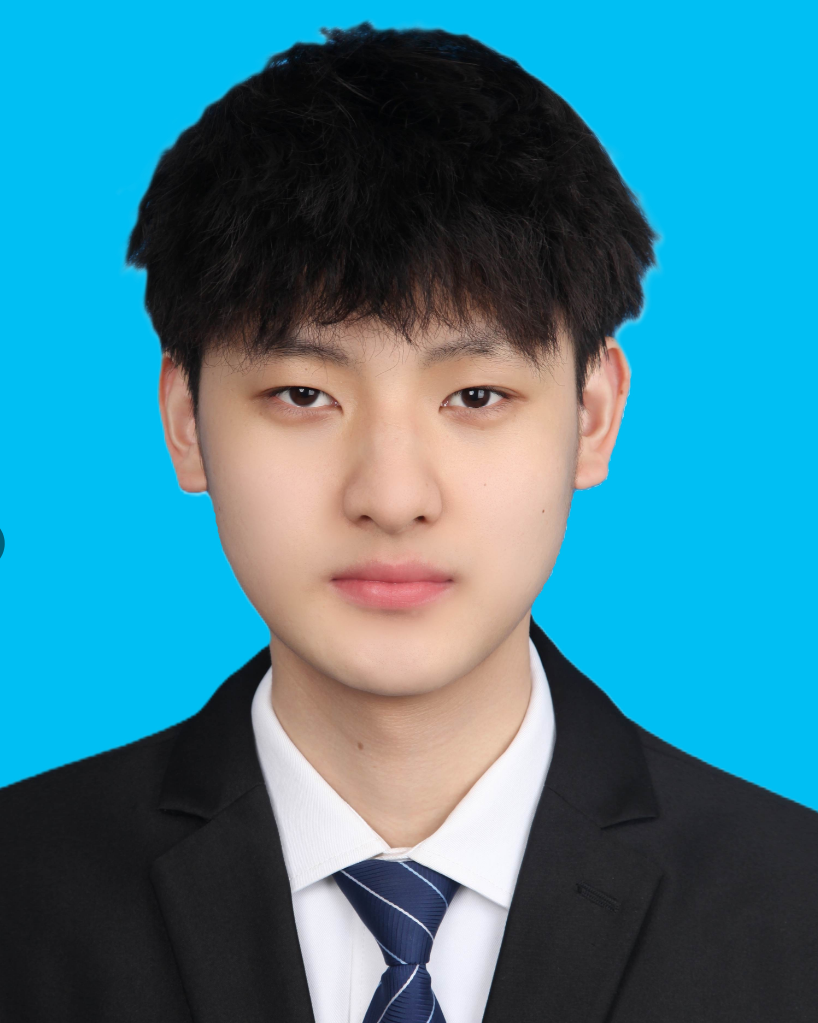}}]{Wei Dai} He is a third-year undergraduate student at the School of Communication Engineering, Xidian University. He is currently a visiting student at the Institute of Data and Information, Tsinghua Shenzhen International Graduate School. His research interests include federated learning and transfer learning.
\end{IEEEbiography}

\vspace{-1.5cm}
\begin{IEEEbiography}[{\includegraphics[width=0.9in,height=1.25in,clip,keepaspectratio]{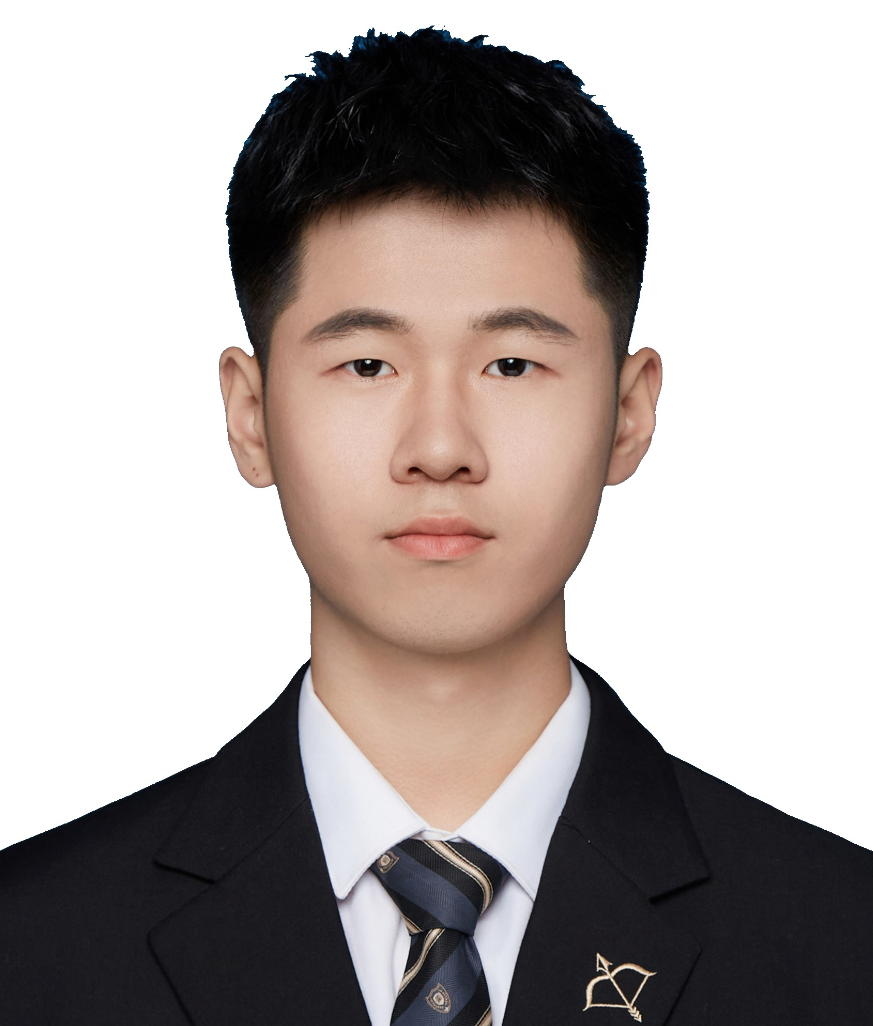}}]{Bowei Liu} is an undergraduate student majoring in Artificial Intelligence at Xidian University, enrolled in 2021. His research interests include long-tailed learning, federated learning, and prompt learning.
\end{IEEEbiography}

\vspace{-1.5cm}
\begin{IEEEbiography}[{\includegraphics[width=0.9in,height=1.25in,clip,keepaspectratio]{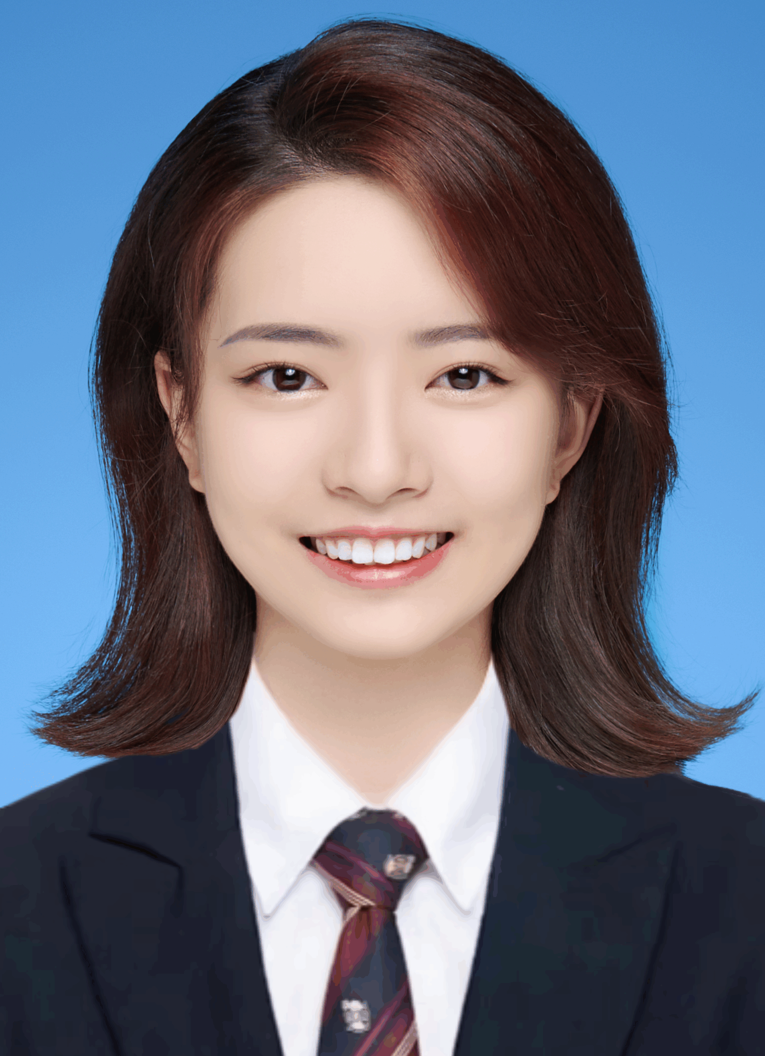}}]{Jiayi Chen}
is currently a third-year undergraduate student in Communication Engineering at Xidian University. She is also pursuing a dual degree in Telecommunications Engineering at Heriot-Watt University.Her research interests include long-tailed learning and semi-supervised learning. 
\end{IEEEbiography}

\vspace{-1.5cm}
\begin{IEEEbiography}[{\includegraphics[width=0.9in,height=1.25in,clip,keepaspectratio]{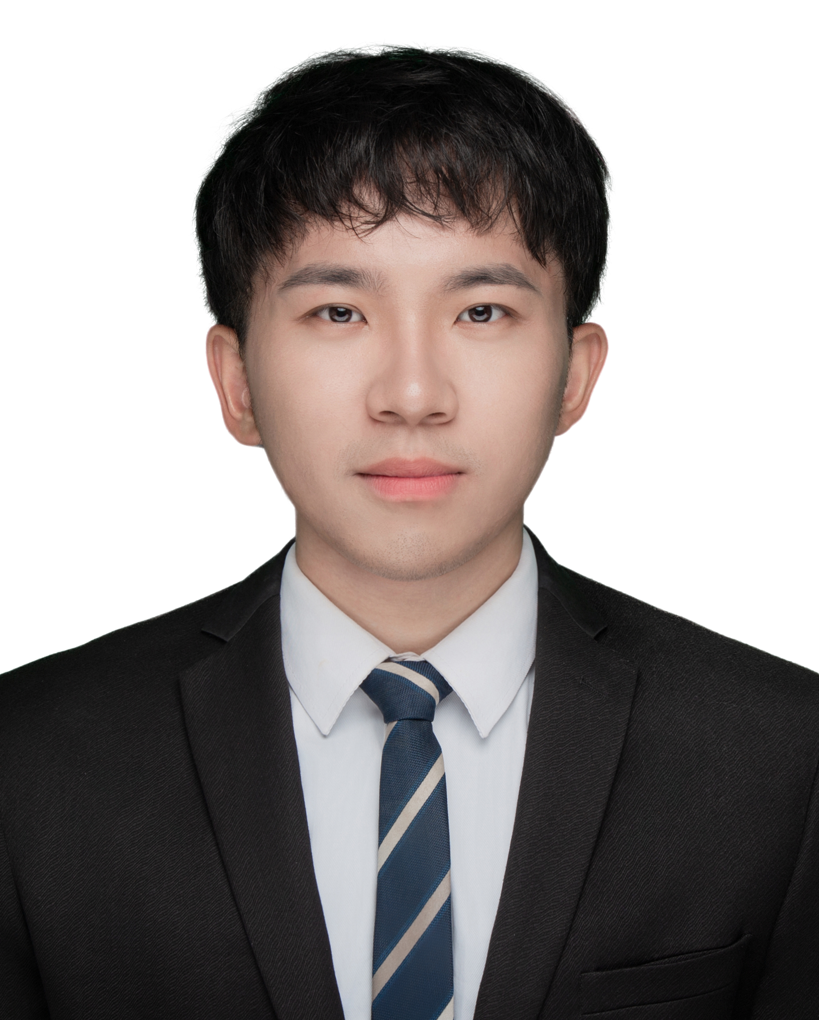}}]{Wenke Huang} received the B.E. degree at the School of Computer Science, Wuhan University, Wuhan, China, in 2021. He is currently a Ph.D. student at the School of Computer Science, Wuhan University, Wuhan, China. His research interests focus on Federated Learning.
\end{IEEEbiography}

\vspace{-1.5cm}
\begin{IEEEbiography}[{\includegraphics[width = 0.9in, height = 1.2in, clip, keepaspectratio]{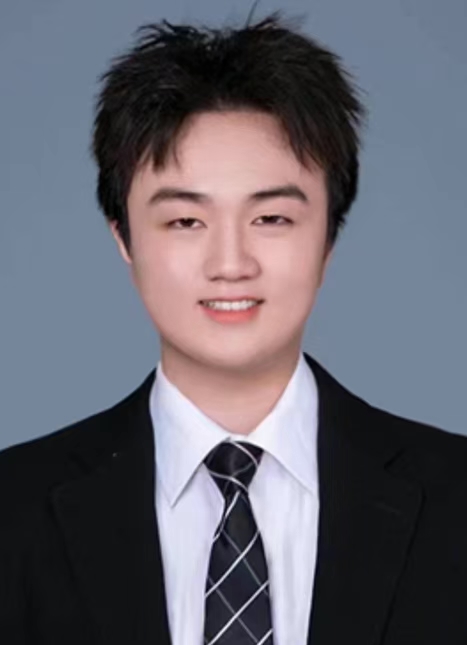}}]{Guancheng Wan} is currently pursuing the Ph.D. degree at the Department of Computer Science, University of California, Los Angeles. His research interests are MLLM and federated learning.
\end{IEEEbiography}

\vspace{-1.5cm}
\begin{IEEEbiography}[{\includegraphics[width=0.9in,height=1.25in,clip,keepaspectratio]{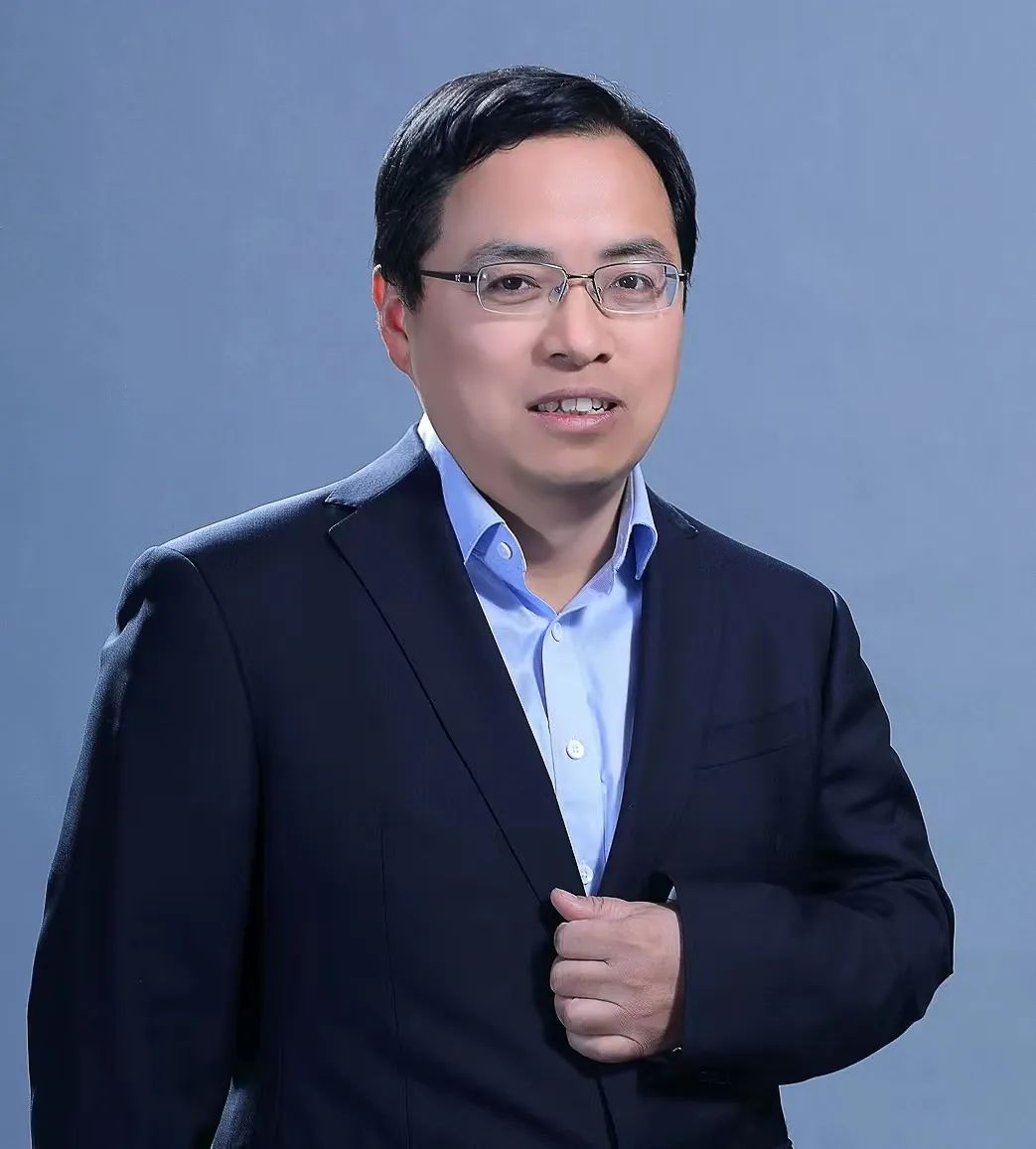}}]{Zhiwu Lu}
is a professor and doctoral supervisor at the Gaoling School of Artificial Intelligence, Renmin University of China. He received his Master of Science degree from the Department of Information Science, School of Mathematical Sciences, Peking University in 2005, and earned his PhD in Computer Science from City University of Hong Kong in 2011. His primary research interests include machine learning, computer vision, and related areas.
\end{IEEEbiography}

\vspace{-1.5cm}
\begin{IEEEbiography}[{\includegraphics[width=0.9in,height=1.25in,clip,keepaspectratio]{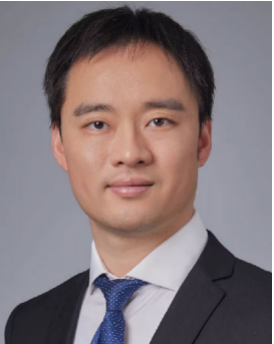}}]{Junchi Yan} is a Professor and Deputy Director with School of Artificial Intelligence, Shanghai Jiao Tong University, Shanghai, China. Before that, he was a full-time Senior Research Staff Member with IBM Research and later an affiliated consultant Senior Researcher with AWS AI Lab. His research interests include machine learning. He is the Associate Editor for IEEE TPAMI, Pattern Recognition etc. He is a Fellow of IAPR, AAIA and IET. He received the IEEE CS AI’10 to Watch Award 2024, IEEE CIS Early Career Award 2025, and CVPR 2024 Best Paper Candidate.
\end{IEEEbiography}



%




\end{document}